\documentclass[10pt,twocolumn,letterpaper]{article}

\usepackage{iccv}
\usepackage{times}
\usepackage{epsfig}
\usepackage{graphicx}
\usepackage{amsmath}
\usepackage{amssymb}

\usepackage{algorithm}
\usepackage{algorithmic}
\usepackage{subfigure}
\usepackage{booktabs} 
\usepackage{multirow}
\usepackage{pifont}
\DeclareMathOperator*{\argmax}{arg\,max}
\DeclareMathOperator*{\argmin}{arg\,min}

\newcommand{\xmark}{\ding{55}}

\newcommand{\citep}[1]{\cite{#1}}
\newcommand{\citet}[1]{\cite{#1}}

\def\arxiv{} 
\def\printappendix{} 
\ifdefined\printappendix
    \def\appendixword{Appendix}
\else
    \def\appendixword{the supplementary material}
\fi

\def\onewidth{.86\linewidth}
\def\twoawidth{.36\linewidth}
\def\twobwidth{.48\linewidth}
\def\twocwidth{.36\linewidth}
\def\twodwidth{.48\linewidth}
\def\aonewidth{.48\linewidth}
\def\conewidth{.36\linewidth}
\def\ctwowidth{.36\linewidth}

\iffalse 
    \newcommand{\kbcomment}[1]{\textcolor{magenta}{kb: #1}}
    \newcommand{\kmcomment}[1]{\textcolor{red}{km: #1}}
    
\else
    \newcommand{\kbcomment}[1]{}
    \newcommand{\kmcomment}[1]{}
    
\fi

\newcommand{\addparagraphup}{\vspace*{2pt}}
\iftrue 
        \newcommand{\cutsectionup}{\vspace*{-2pt}}
        \newcommand{\cutsectiondown}{\vspace*{-2pt}}

        \newcommand{\cutsubsectionup}{\vspace*{-2pt}}
        \newcommand{\cutsubsectiondown}{\vspace*{-2pt}}

        \newcommand{\cuthalfcaptionup}{\vspace*{-4pt}}
        \newcommand{\cuthalfcaptiondown}{\vspace*{-4pt}}

        \newcommand{\cutcaptionup}{\vspace*{-4pt}}
        \newcommand{\cutcaptiondown}{\vspace*{-12pt}}

        \newcommand{\cuthalftablecaptionup}{\vspace*{-0pt}}
        \newcommand{\cuthalftablecaptiondown}{\vspace*{-0pt}}

        \newcommand{\cuttablecaptionup}{\vspace*{-4pt}}
        \newcommand{\cuttablecaptiondown}{\vspace*{-0pt}}

        \newcommand{\cutequationup}{\vspace*{-0pt}}
        \newcommand{\cutequationdown}{\vspace*{-0pt}}

        \newcommand{\cuthalftableup}{\vspace*{-0pt}}
        \newcommand{\cuthalftabledown}{\vspace*{-0pt}}
        
        \newcommand{\cuttableup}{\vspace*{-0pt}}
        \newcommand{\cuttabledown}{\vspace*{-14pt}}
        
        \newcommand{\cutabstractup}{\vspace*{-10pt}}
        \newcommand{\cutabstractdown}{\vspace*{-14pt}}

\else 
        \newcommand{\cutsectionup}{}
        \newcommand{\cutsectiondown}{}

        \newcommand{\cutsubsectionup}{}
        \newcommand{\cutsubsectiondown}{}

        \newcommand{\cuthalfcaptionup}{}
        \newcommand{\cuthalfcaptiondown}{}

        \newcommand{\cutcaptionup}{}
        \newcommand{\cutcaptiondown}{}

        \newcommand{\cuthalftablecaptionup}{}
        \newcommand{\cuthalftablecaptiondown}{}

        \newcommand{\cuttablecaptionup}{}
        \newcommand{\cuttablecaptiondown}{}

        \newcommand{\cutequationup}{}
        \newcommand{\cutequationdown}{}

        \newcommand{\cuthalftableup}{}
        \newcommand{\cuthalftabledown}{}
        
        \newcommand{\cuttableup}{}
        \newcommand{\cuttabledown}{}

        \newcommand{\cutabstractup}{}
        \newcommand{\cutabstractdown}{}

\fi

\ifdefined\arxiv 
\usepackage[pagebackref=true,breaklinks=true,letterpaper=true,colorlinks,bookmarks=false]{hyperref}
\else 
\usepackage[breaklinks=true,bookmarks=false]{hyperref} 
\fi 

\iccvfinalcopy 

\def\iccvPaperID{3510} 

\ifdefined\arxiv 
\else 
\ificcvfinal\fi
\fi 

\begin{document}

%
%
\title{Overcoming Catastrophic Forgetting with Unlabeled Data in the Wild}

\author{Kibok Lee$^{*}$\quad Kimin Lee$^{\dagger}$\quad Jinwoo Shin$^{\dagger}$\quad Honglak Lee$^{*}$\\
$^{*}$University of Michigan\quad 
$^{\dagger}$KAIST\\
$^{*}${\tt\small \{kibok, honglak\}@umich.edu}\quad 
$^{\dagger}${\tt\small \{kiminlee, jinwoos\}@kaist.ac.kr}
}

\maketitle
\ifdefined\arxiv 
\else 
\ificcvfinal\thispagestyle{empty}\fi
\fi 

\begin{abstract}
\cutabstractup
Lifelong learning with deep neural networks is well-known to suffer from catastrophic forgetting:
the performance on previous tasks drastically degrades when learning a new task.
To alleviate this effect, we propose to leverage 
a large stream of unlabeled data
easily obtainable
in the wild.
In particular,
we design
a novel class-incremental learning scheme with
(a)~a new distillation loss, termed global distillation,
(b)~a learning strategy to avoid overfitting to the most recent task, and
(c)~a confidence-based sampling method
to 
effectively 
leverage unlabeled external data.
Our experimental results on various datasets, including CIFAR and ImageNet,
demonstrate the superiority of the proposed methods over prior methods, particularly when a stream of unlabeled data is accessible:
our method
shows up to 15.8\% higher accuracy and 46.5\% less forgetting 
compared to the state-of-the-art method.
The code is available at
\textcolor{magenta}{\url{https://github.com/kibok90/iccv2019-inc}}.
\end{abstract}
\cutabstractdown

\cutsectionup
\section{Introduction} \label{sec:intro}
\cutsectiondown

Deep neural networks (DNNs) have achieved remarkable success in many machine learning applications,
e.g., classification \cite{he2016deep},
generation \cite{miyato2018spectral},
object detection \cite{he2017mask}, and
reinforcement learning \cite{silver2017mastering}.
However, in the real world where the number of tasks continues to grow, the entire tasks cannot be given at once;
rather, it may be given as a sequence of tasks.
The goal of class-incremental learning \cite{rebuffi2017icarl} is to enrich the ability of a model dealing with such a case, by aiming to
perform both previous and new tasks well.\footnote{In class-incremental learning, a set of classes is given in each task.
In evaluation, it aims to classify data in any class learned so far without task boundaries.}
In particular, it has gained much attention recently as DNNs tend to forget previous tasks easily when learning new tasks,
which is a phenomenon called catastrophic forgetting \cite{french1999catastrophic, mccloskey1989catastrophic}.

The primary reason of catastrophic forgetting is the limited resources for scalability:
all training data of previous tasks cannot be stored in a limited size of memory 
as the number of tasks increases.
Prior works in class-incremental learning focused on learning in a closed environment, i.e., a model can only see
the given labeled training dataset during training
\cite{castro2018end, hou2018lifelong, li2018supportnet, li2016learning, rebuffi2017icarl}.
However, in the real world,
we live with a continuous and large stream of 
unlabeled data 
easily obtainable on the fly or
transiently, e.g., by data mining on social media \cite{mahajan2018exploring} and web data \cite{krause2016unreasonable}.
Motivated by this, we propose to leverage such a large stream of unlabeled external data for overcoming
catastrophic forgetting.
We remark that our setup on unlabeled data
is similar to self-taught learning \citet{raina2007self} rather than semi-supervised learning,
because we do not assume any correlation between
unlabeled data and the labeled training dataset.


\begin{figure}[t]
\centering
\includegraphics[width=\onewidth]{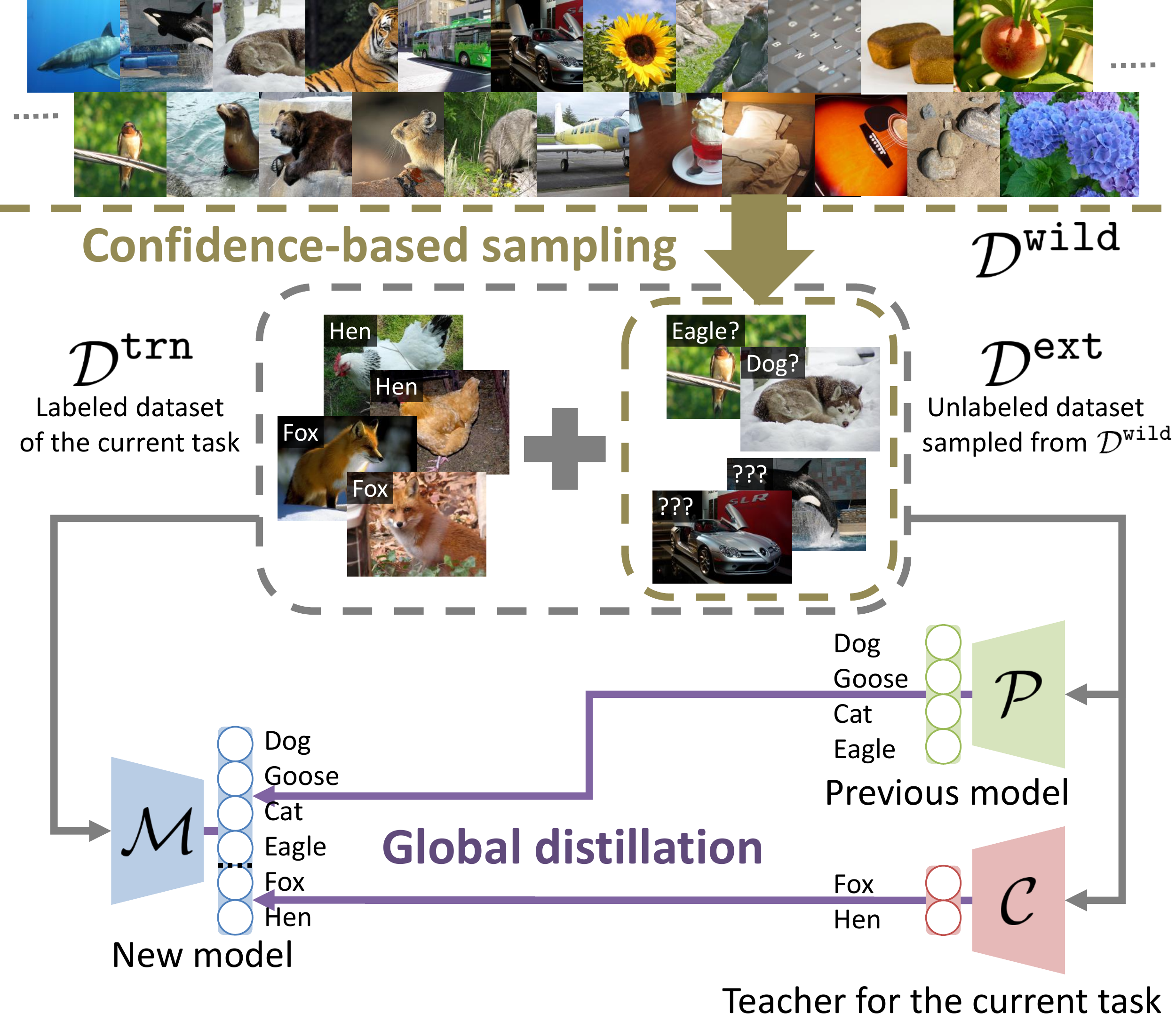}
\cuthalfcaptionup
\caption{
We propose to leverage a large stream of unlabeled data in the wild for class-incremental learning.
At each stage, a confidence-based sampling strategy is applied to build an external dataset.
Specifically, some of unlabeled data are sampled based on the prediction of the model learned in the previous stage $\mathcal{P}$ for alleviating catastrophic forgetting, and some of them are randomly sampled for confidence calibration.
Under the combination of the labeled training dataset and the unlabeled external dataset, a teacher model $\mathcal{C}$ first learns the current task, and then the new model $\mathcal{M}$ learns both the previous and current tasks by distilling the knowledge of $\mathcal{P}$, $\mathcal{C}$, and their ensemble $\mathcal{Q}$.
}
\cuthalfcaptiondown
\label{fig:overview}
\end{figure}


\addparagraphup\noindent{\bf Contribution.}
Under the new class-incremental setup, our contribution is three-fold (see Figure~\ref{fig:overview} for an overview):


\begin{itemize}
    \item[$\mathcal A.$]
    We propose a new 
    learning objective,
    termed global distillation, which utilizes data to distill the knowledge of
    reference models
    effectively.
    \item[$\mathcal B$.] We design a 3-step learning scheme to improve the effectiveness of global distillation:
    (i)~training a teacher specialized for the current task,
    (ii)~training a model by distilling the knowledge of the previous model, the teacher learned in (i), and their ensemble, and
    (iii)~fine-tuning to avoid overfitting to the current task.
    \item[$\mathcal C$.] We propose
    a confidence-based sampling method
    to effectively leverage a large stream of unlabeled data.
\end{itemize}


In the contribution $\mathcal A$,
global distillation encourages the model to learn knowledge over all previous tasks, while prior works only applied a task-wise local distillation \cite{castro2018end, hou2018lifelong, li2016learning, rebuffi2017icarl}.
In particular, the proposed global distillation
distills the knowledge of how to distinguish
classes across different tasks, 
while local distillation does not. 
We show that the performance gain due to global distillation is particularly significant if some unlabeled external data are available.

In the contribution $\mathcal B$, the first two steps (i), (ii) of the proposed learning scheme are designed to keep the knowledge of the previous tasks, as well as to learn the current task.
On the other hand, the purpose of the last step (iii)
is to avoid overfitting to the current task:
due to the scalability issue, only a small portion of data in the previous tasks are kept and replayed during training \cite{castro2018end, nguyen2018variational, rebuffi2017icarl}.
This inevitably incurs bias in the prediction of the learned model, being favorable for the current task.
To mitigate the issue of imbalanced training, we fine-tune the model based on the statistics of data in the previous and current tasks.

Finally, the contribution $\mathcal C$ is motivated from the intuition that
as the data distribution of unlabeled data is more similar to that of the previous tasks, it prevents the model from catastrophic forgetting more.
Since unlabeled data in the wild is not necessarily related to the previous tasks,
it is far from being clear whether they contain an useful information for alleviating catastrophic forgetting.
Therefore, we propose to sample an external dataset by a principled sampling strategy.
To sample an effective external dataset from a large stream of unlabeled data, we propose to train a confidence-calibrated model \cite{lee2018training, lee2018hierarchical} by utilizing irrelevant data as out-of-distribution (OOD)\footnote{Out-of-distribution refers to the data distribution being far from those of the tasks learned so far.} samples.
We show that unlabeled data from OOD should also be sampled for maintaining the model to be more confidence-calibrated.

Our experimental results demonstrate the superiority of the proposed methods over prior methods.
In particular,
we show that the performance gain in the proposed methods is more significant
when unlabeled external data are available.
For example, under our experiment setup
on ImageNet \cite{deng2009imagenet},
our method with an external dataset achieves 15.8\% higher accuracy and
46.5\% less forgetting 
compared to the state-of-the-art method (E2E) \cite{castro2018end} (4.8\% higher accuracy and 6.0\% less forgetting without an external dataset).



\cutsectionup
\section{Approach} \label{sec:approach}
\cutsectiondown

In this section, we propose a new learning method for class-incremental learning.
In Section \ref{sec:prelim}, we further describe the scenario and
learning objectives.
In Section~\ref{sec:gd}, we propose a novel learning objective, termed global distillation.
In Section~\ref{sec:sample}, we propose a confidence-based sampling strategy to build an external dataset from a large stream of unlabeled data.

\cutsubsectionup
\subsection{Preliminaries: Class-Incremental Learning}\label{sec:prelim}
\cutsubsectiondown

Formally, let $(x,y) \in \mathcal{D}$ be a data $x$ and its label $y$ in a dataset $\mathcal{D}$, and 
let $\mathcal{T}$ be a supervised task mapping $x$ to $y$.
We denote $y \in \mathcal{T}$ if $y$ is in the range of $\mathcal{T}$
such that $\lvert \mathcal{T} \rvert$ is the number of class labels in $\mathcal{T}$.
For the $t$-th task $\mathcal{T}_{t}$, let $\mathcal{D}_{t}$ be the corresponding training dataset, and
$\mathcal{D}_{t-1}^{\mathtt{cor}} \subseteq \mathcal{D}_{t-1} \cup \mathcal{D}_{t-2}^{\mathtt{cor}}$ be a coreset\footnote{Coreset is a small dataset kept in a limited amount of memory used to replay previous tasks.
Initially, $\mathcal{D}_{0}^{\mathtt{cor}} = \emptyset$.}
containing representative data of previous tasks $\mathcal{T}_{1:(t-1)} = \{ \mathcal{T}_{1}, \dots, \mathcal{T}_{t-1} \}$,
such that $\mathcal{D}_{t}^{\mathtt{trn}} = \mathcal{D}_{t} \cup \mathcal{D}_{t-1}^{\mathtt{cor}}$ is the entire labeled training dataset available at the $t$-th stage.
Let $\mathcal{M}_{t} = \{ \theta, \phi_{1:t} \}$ be the set of learnable parameters of a model, where $\theta$ 
and $\phi_{1:t} = \{ \phi_{1}, \dots, \phi_{t} \}$ indicate shared 
and task-specific parameters, respectively.\footnote{If multiple task-specific parameters are given,
then logits of all classes are concatenated for prediction without task boundaries.
Note that tasks do not have to be disjoint, such that a class can appear in multiple tasks.}

The goal at the $t$-th stage is to train a model $\mathcal{M}_{t}$ to
perform the current task $\mathcal{T}_{t}$ as well as the previous tasks $\mathcal{T}_{1:(t-1)}$ without task boundaries, i.e., all class labels in $\mathcal{T}_{1:t}$ are candidates at test time.
To this end,
a small coreset $\mathcal{D}_{t-1}^{\mathtt{cor}}$
and the previous model $\mathcal{M}_{t-1}$ are transferred from the previous stage.
We also assume that a large stream of unlabeled data is accessible, and
an essential external dataset $\mathcal{D}_{t}^{\mathtt{ext}}$ is sampled, where the sampling method is described in Section~\ref{sec:sample}.
Note that we do not assume any correlation between the stream of unlabeled data and the tasks.
The outcome at the $t$-th stage is the model $\mathcal{M}_{t}$ that can perform all observed tasks $\mathcal{T}_{1:t}$, and the coreset $\mathcal{D}_{t}^{\mathtt{cor}}$ for learning in subsequent stages.

\addparagraphup\noindent{\bf Learning objectives.}
When a dataset $\mathcal{D}$ is labeled,
the standard way of training a classification model $\mathcal{M} = \{ \theta, \phi \}$ is to optimize the cross-entropy loss:
\cutequationup
\begin{align*}
&\mathcal{L}_{\mathtt{cls}}(\theta, \phi; \mathcal{D}) =
\frac1{\lvert \mathcal{D} \rvert} \sum_{(x,y) \in \mathcal{D}}
[ -\log p(y | x ; \theta, \phi) ].
\end{align*}
\cutequationdown
On the other hand, if we have a reference model $\mathcal{R} = \{\theta^{\mathcal{R}}, \phi^{\mathcal{R}}\}$,
the dataset $\mathcal{D}$ does not require any label because the target label is given by $\mathcal{R}$:
\cutequationup
\begin{align*}
&\mathcal{L}_{\mathtt{dst}}(\theta, \phi; \mathcal{R}, \mathcal{D}) \\
&\qquad=
\frac1{\lvert \mathcal{D} \rvert} \sum_{x \in \mathcal{D}} \sum_{y \in \mathcal{T}}
[ - p(y | x ; \theta^{\mathcal{R}}, \phi^{\mathcal{R}}) \log p(y | x ; \theta, \phi) ],
\end{align*}
\cutequationdown
where the probabilities can be smoothed for better distillation (see \citet{hinton2015distilling} or \appendixword).

\addparagraphup\noindent{\bf Previous approaches.}
At the $t$-th stage, the standard approach to train a model $\mathcal{M}_{t}$ is to minimize the following classification loss:
\cutequationup
\begin{align}
\mathcal{L}_{\mathtt{cls}}(\theta, \phi_{1:t}; \mathcal{D}_{t}^{\mathtt{trn}}).
\label{eq:gc}
\end{align}
\cutequationdown
However, in class-incremental learning, the limited size of the coreset makes the learned model suffer from catastrophic forgetting.
To overcome this, the previous model $\mathcal{P}_{t} = \{ \theta^{\mathcal{P}}, \phi_{1:(t-1)}^{\mathcal{P}} \} \triangleq \mathcal{M}_{t-1}$ has been utilized to generate soft labels, which is the knowledge of $\mathcal{P}_{t}$ about the given data \cite{castro2018end, hou2018lifelong, li2016learning, rebuffi2017icarl}:
\cutequationup
\begin{align}
\sum_{s=1}^{t-1} \mathcal{L}_{\mathtt{dst}}(\theta, \phi_{s}; \mathcal{P}_{t}, \mathcal{D}_{t}^{\mathtt{trn}}),
\label{eq:ld}
\end{align}
\cutequationdown
where this objective is jointly optimized with Eq.~\eqref{eq:gc}.
We call this task-wise knowledge distillation as local distillation (LD), which transfers the knowledge within each of the tasks.
However, because they are defined in a task-wise manner, this objective misses the knowledge about discrimination between classes in different tasks.

\cutsubsectionup
\subsection{Global Distillation} \label{sec:gd}
\cutsubsectiondown

Motivated by the limitation of LD, we propose to distill the knowledge of reference models \emph{globally}.
With the reference model $\mathcal{P}_{t}$, the knowledge can be globally distilled by minimizing the following loss:
\cutequationup
\begin{align}
\mathcal{L}_{\mathtt{dst}}(\theta, \phi_{1:(t-1)}; \mathcal{P}_{t}, \mathcal{D}_{t}^{\mathtt{trn}} \cup \mathcal{D}_{t}^{\mathtt{ext}}).
\label{eq:gd_p}
\end{align}
\cutequationdown
However, learning by minimizing Eq.~\eqref{eq:gd_p} would cause a bias:
since $\mathcal{P}_{t}$ did not learn to perform the current task $\mathcal{T}_{t}$,
the knowledge about the current task would not be properly learned when only Eq.~\eqref{eq:gc}+\eqref{eq:gd_p} are minimized,
i.e., the performance on the current task would be unnecessarily sacrificed.
To compensate for this, we introduce another teacher model $\mathcal{C}_{t} = \{ \theta^{\mathcal{C}}, \phi_{t}^{\mathcal{C}} \}$ specialized for the current task $\mathcal{T}_{t}$:
\cutequationup
\begin{align}
\mathcal{L}_{\mathtt{dst}}(\theta, \phi_{t}; \mathcal{C}_{t}, \mathcal{D}_{t}^{\mathtt{trn}} \cup \mathcal{D}_{t}^{\mathtt{ext}}).
\label{eq:gd_c}
\end{align}
\cutequationdown
This model can be trained by minimizing the standard cross-entropy loss:
\cutequationup
\begin{align}
\mathcal{L}_{\mathtt{cls}}(\theta^\mathcal{C}, \phi_{t}^\mathcal{C}; \mathcal{D}_{t}).
\label{eq:tc}
\end{align}
\cutequationdown
Note that only the dataset of the current task $\mathcal{D}_{t}$ is used, because $\mathcal{C}_{t}$ is specialized for the current task only.
We revise this loss in Section~\ref{sec:sample} for better external data sampling.

However, as $\mathcal{P}_{t}$ and $\mathcal{C}_{t}$ learned to perform only $\mathcal{T}_{1:(t-1)}$ and $\mathcal{T}_{t}$, respectively,
discrimination between $\mathcal{T}_{1:(t-1)}$ and $\mathcal{T}_{t}$ is not possible with the knowledge distilled from these two reference models.
To complete the missing knowledge, we define $\mathcal{Q}_{t}$ as an ensemble of $\mathcal{P}_{t}$ and $\mathcal{C}_{t}$:
let
\begin{align*}
p_{\max} &= \max_{y} p(y | x, \theta^{\mathcal{P}}, \phi_{1:(t-1)}^{\mathcal{P}}), \\
y_{\max} &= \argmax_{y} p(y | x, \theta^{\mathcal{P}}, \phi_{1:(t-1)}^{\mathcal{P}}).
\end{align*}
Then, the output of $\mathcal{Q}_{t}$ can be defined as:
\cutequationup
\begin{align}
&p(y | x, \theta^{\mathcal{Q}}, \phi_{1:t}^{\mathcal{Q}}) =
\nonumber \\
&\begin{cases}
p_{\max} & \text{if } y = y_{\max}, \\
\frac{1 - p_{\max} - \varepsilon}{1 - p_{\max}} p(y | x, \theta^{\mathcal{P}}, \phi_{1:(t-1)}^{\mathcal{P}}) & \text{elif } y \in \mathcal{T}_{1:(t-1)},
\\
\varepsilon p(y | x, \theta^{\mathcal{C}}, \phi_{t}^{\mathcal{C}}) & \text{elif } y \in \mathcal{T}_{t},
\end{cases}
\label{eq:ensemble}
\end{align}
\cutequationdown
such that $\sum_{y} p(y | x, \theta^{\mathcal{Q}}, \phi_{1:t}^{\mathcal{Q}}) = 1$.
Here, $\varepsilon$ adjusts the confidence about whether the given data is in $\mathcal{T}_{1:(t-1)}$ or $\mathcal{T}_{t}$.
This information is basically missing,
however, can be computed with an assumption that the expected predicted probability is the same over all negative classes $\forall y \neq y_{\max}$,
i.e.,
$\mathbb{E}_{y} \left[ p_{\varepsilon} (y | x, \theta^{\mathcal{P}}, \phi_{1:(t-1)}^{\mathcal{P}}) \right] = \mathbb{E}_{y \neq  y_{\max}} \left[ p_{\varepsilon} (y | x, \theta^{\mathcal{C}}, \phi_{t}^{\mathcal{C}}) \right]$:
\cutequationup
\begin{align}
\varepsilon = \frac{(1 - p_{\max}) \lvert \mathcal{T}_{t} \rvert}{\lvert \mathcal{T}_{1:t} \rvert - 1}.
\label{eq:eps}
\end{align}
\cutequationdown
Since the ensemble model $\mathcal{Q}_{t}$ is able to perform all tasks, all parameters can be updated:
\cutequationup
\begin{align}
\mathcal{L}_{\mathtt{dst}}(\theta, \phi_{1:t}; \mathcal{Q}_{t}, \mathcal{D}_{t}^{\mathtt{ext}}).
\label{eq:gd_q}
\end{align}
\cutequationdown
Note that the labeled dataset $\mathcal{D}_{t}^{\mathtt{trn}}$ is not used,
because 
it is already used in Eq.~\eqref{eq:gc} for the same range of classes.

Finally, our global distillation (GD) model learns by optimizing Eq.~\eqref{eq:gc}+\eqref{eq:gd_p}+\eqref{eq:gd_c}+\eqref{eq:gd_q}:
\cutequationup
\begin{align}
\mathcal{L}_{\mathtt{cls}}(\theta, \phi_{1:t}; \mathcal{D}_{t}^{\mathtt{trn}})
&+ \mathcal{L}_{\mathtt{dst}}(\theta, \phi_{1:(t-1)}; \mathcal{P}_{t}, \mathcal{D}_{t}^{\mathtt{trn}} \cup \mathcal{D}_{t}^{\mathtt{ext}})
\nonumber \\
&+ \mathcal{L}_{\mathtt{dst}}(\theta, \phi_{t}; \mathcal{C}_{t}, \mathcal{D}_{t}^{\mathtt{trn}} \cup \mathcal{D}_{t}^{\mathtt{ext}})
\nonumber \\
&+ \mathcal{L}_{\mathtt{dst}}(\theta, \phi_{1:t}; \mathcal{Q}_{t}, \mathcal{D}_{t}^{\mathtt{ext}}).
\label{eq:gd}
\end{align}
\cutequationdown
We study the contribution of each
term in Table~\ref{tb:loss}.

\addparagraphup\noindent{\bf Balanced fine-tuning.}
The statistics of class labels in the training dataset is also an information learned during training.
Since the number of data from the previous tasks is much smaller than that of the current task,
the prediction of the model is biased to the current task.
To remove the bias, we further fine-tune the model after training with the same learning objectives.
When fine-tuning, for each loss with $\mathcal{D}$ and $\mathcal{T}$,
we scale the gradient computed from a data with label $k \in \mathcal{T}$ by the following:
\cutequationup
\begin{align}
w_{\mathcal{D}}^{(k)} \propto
\frac1{\lvert \{ (x,y) \in \mathcal{D} | y=k \} \rvert}.
\label{eq:dw}
\end{align}
\cutequationdown
Since scaling a gradient is equivalent to feeding the same data multiple times,
we call this method \emph{data weighting}.
We also normalize the weights by multiplying them with $\lvert \mathcal{D} \rvert / \lvert \mathcal{T} \rvert$, such that they are all one if $\mathcal{D}$ is balanced.

We only fine-tune the task-specific parameters $\phi_{1:t}$ with data weighting,
because all training data would be equally useful for representation learning, i.e., shared parameters $\theta$,
while the bias in the data distribution of the training dataset should be removed when training a classifier, i.e., $\phi_{1:t}$.
The effect of balanced fine-tuning can be found in Table~\ref{tb:balance}.

\addparagraphup\noindent{\bf Loss weight.}
We balance the contribution of each loss by the relative size of each task learned in the loss:
for each loss for learning $\mathcal{T}$, the loss weight at the $t$-th stage is

\cutequationup
\begin{align}
w^{L} =
\frac{\lvert \mathcal{T} \rvert}{\lvert \mathcal{T}_{1:t} \rvert}.
\label{eq:lw}
\end{align}
\cutequationdown

We note that the loss weight can be tuned as
a hyperparameter, but we find that this loss weight performs better than other values in general, as it follows the statistics of the test dataset:
all classes are equally likely to be appeared.


\begin{algorithm}[t]
  \caption{3-step learning with GD.}
  \label{alg:learning}
\begin{algorithmic}[1]
  \STATE $t = 1$
  \WHILE{{\bfseries true}}
    \STATE {\bfseries Input:} previous model $\mathcal{P}_{t} = \mathcal{M}_{t-1}$, coreset $\mathcal{D}_{t-1}^{\mathtt{cor}}$, \\ \hspace*{10pt} training dataset $\mathcal{D}_{t}$, unlabeled data stream $\mathcal{D}_{t}^{\texttt{wild}}$
    \STATE {\bfseries Output:} new coreset $\mathcal{D}_{t}^{\mathtt{cor}}$, model $\mathcal{M}_{t} = \{ \theta, \phi_{1:t} \}$
    \STATE $\mathcal{D}_{t}^{\mathtt{trn}} = \mathcal{D}_{t} \cup \mathcal{D}_{t-1}^{\mathtt{cor}}$
    \STATE $N_{C} = \lvert \mathcal{D}_{t-1}^{\mathtt{cor}} \rvert$, $N_{D} = \lvert \mathcal{D}_{t}^{\mathtt{trn}} \rvert$
    \STATE Sample $\mathcal{D}_{t}^{\mathtt{ext}}$ from $\mathcal{D}_{t}^{\texttt{wild}}$ using Algorithm~\ref{alg:sample}
    \STATE Train $\mathcal{C}_{t}$ by minimizing Eq.~\eqref{eq:tc_ex}
    \IF{$t > 1$}
      \STATE Train $\mathcal{M}_{t}$ by minimizing
       Eq.~\eqref{eq:gd}
      \STATE Fine-tune $\phi_{1:t}$ by minimizing
      Eq.~\eqref{eq:gd}, \\
      with data weighting in Eq.~\eqref{eq:dw}
    \ELSE
      \STATE $\mathcal{M}_{t} = \mathcal{C}_{t}$
    \ENDIF
    \STATE Randomly sample $\mathcal{D}_{t}^{\mathtt{cor}} \subseteq \mathcal{D}_{t}^{\mathtt{trn}}$ such that \\ $\lvert \{ (x,y) \in \mathcal{D}_{t}^{\mathtt{cor}} | y=k \} \rvert = N_{C} / \lvert \mathcal{T}_{1:t} \rvert$ for $k \in \mathcal{T}_{1:t}$
    \STATE $t=t+1$
  \ENDWHILE
\end{algorithmic}
\end{algorithm}

\addparagraphup\noindent{\bf 3-step learning algorithm.}
In summary, our learning strategy has three steps:
training $\mathcal{C}_{t}$ specialized for the current task $\mathcal{T}_{t}$,
training $\mathcal{M}_{t}$
by distilling the knowledge of the reference models $\mathcal{P}_{t}$, $\mathcal{C}_{t}$, and $\mathcal{Q}_{t}$, and
fine-tuning the task-specific parameters $\phi_{1:t}$ with data weighting.
Algorithm~\ref{alg:learning} describes the 3-step learning scheme.

For coreset management, we build a balanced coreset by randomly selecting data for each class.
We note that other more sophisticated selection algorithms like herding \cite{rebuffi2017icarl} 
do not perform significantly better than random selection, which is also reported in prior works \cite{castro2018end, wu2018incremental}.

\begin{algorithm}[t]
  \caption{Sampling external dataset.}
  \label{alg:sample}
\begin{algorithmic}[1]
  \STATE {\bfseries Input:} previous model $\mathcal{P}_{t} = \{ \theta^{\mathcal{P}}, \phi_{1:(t-1)}^{\mathcal{P}} \}$, \\ \hspace*{28pt} unlabeled data stream $\mathcal{D}_{t}^{\texttt{wild}}$, sample size $N_{D}$, \\ \hspace*{28pt} number of unlabeled data to be retrieved $N_{\max}$
    \STATE {\bfseries Output:} sampled external dataset $\mathcal{D}_{t}^{\mathtt{ext}}$
    \STATE $\mathcal{D}^{\text{prev}} = \emptyset$, $\mathcal{D}^{\text{OOD}} = \emptyset$
    \STATE $N_{\text{prev}} = 0.3 N_{D}$, $N_{\text{OOD}} = 0.7 N_{D}$
    \STATE $N(k) \triangleq \lvert \{ (x,y,p) \in \mathcal{D}^{\text{prev}} | y = k \} \rvert$
    \WHILE{$\lvert \mathcal{D}^{\text{OOD}} \rvert < N_{\text{OOD}}$}
        \STATE Get $x \in \mathcal{D}_{t}^{\texttt{wild}}$ and update $\mathcal{D}^{\text{OOD}} = \mathcal{D}^{\text{OOD}} \cup \{x\}$
    \ENDWHILE
    \STATE $N_{\text{ret}} = N_{\text{OOD}}$
    \WHILE{$N_{\text{ret}} < N_{\max}$}
        \STATE Get $x \in \mathcal{D}_{t}^{\texttt{wild}}$ and compute the prediction of $\mathcal{P}$: \\
        \hspace*{1pt} $\hat{p} = \max_{y} p \left( y | x; \theta^{\mathcal{P}}, \phi_{1:(t-1)}^{\mathcal{P}} \right)$, \\
        \hspace*{1pt} $\hat{y} = \argmax_{y} p \left( y | x; \theta^{\mathcal{P}}, \phi_{1:(t-1)}^{\mathcal{P}} \right)$
        \IF{$N(\hat{y}) < N_{\text{prev}} / \lvert \mathcal{T}_{1:(t-1)} \rvert$}
            \STATE $\mathcal{D}^{\text{prev}} = \mathcal{D}^{\text{prev}} \cup \{(x,\hat{y},\hat{p})\}$
        \ELSE
            \STATE Replace the least probable data in class $\hat{y}$: \\
            \hspace*{1pt} $(x',\hat{y},p') = \argmin_{\{ (x,y,p) \in \mathcal{D}^{\text{prev}} | y = \hat{y} \}} p$
            \IF{$p' < \hat{p}$}
                \STATE $\mathcal{D}^{\text{prev}} = \left( \mathcal{D}^{\text{prev}} \backslash \{(x',\hat{y},p')\} \right) \cup \{(x,\hat{y},\hat{p})\}$
            \ENDIF
        \ENDIF
        \STATE  $N_{\text{ret}} = N_{\text{ret}} + 1$
    \ENDWHILE
    \STATE {\bf Return} $\mathcal{D}_{t}^{\mathtt{ext}} = \mathcal{D}^{\text{OOD}} \cup \{ x | (x,y,p) \in \mathcal{D}^{\text{prev}} \}$
\end{algorithmic}
\end{algorithm}

\cutsubsectionup
\subsection{Sampling External Dataset} \label{sec:sample}
\cutsubsectiondown

Although a large amount of unlabeled data would be easily obtainable, there are two issues when using them for knowledge distillation:
(a)~training on a large-scale external dataset is expensive, and
(b)~most of the data would not be helpful, because they would be irrelevant to the tasks the model learns.
To overcome these issues, we propose to sample an external dataset useful for knowledge distillation from a large stream of unlabeled data.
Note that the sampled external dataset does not require an additional permanent memory; it is discarded after learning.

\addparagraphup\noindent{\bf Sampling for confidence calibration.}
In order to alleviate catastrophic forgetting caused by the imbalanced training dataset, sampling external data that are expected to be in the previous tasks is desirable.
Since the previous model $\mathcal{P}$ is expected to produce an output with high confidence if the data is likely to be in the previous tasks, the output of $\mathcal{P}$ can be used for sampling.
However, modern DNNs are highly overconfident \cite{guo2017calibration, lee2018training}, thus a model learned with a discriminative loss
would produce a prediction with high confidence even if the data is not from any of the previous tasks.
Since most of the unlabeled data would not be relevant to any of the previous tasks, i.e., they are considered to be from out-of-distribution (OOD),
it is important to avoid overconfident prediction on such irrelevant data.
To achieve this, the model should learn to be confidence-calibrated by learning with a certain amount of OOD data as well as data of the previous tasks \cite{lee2018training, lee2018hierarchical}.
When sampling OOD data, we propose to randomly sample data rather than relying on the confidence of the previous model, as OOD is widely distributed over the data space.
The effect of this sampling strategy can be found in Table~\ref{tb:sample}.
Algorithm~\ref{alg:sample} describes our sampling strategy.
The ratio of OOD data ($N_{\text{prev}} : N_{\text{OOD}}$) is determined by validation; for more details, see \appendixword.
This sampling algorithm can take a long time, but we limit the number of retrieved unlabeled data in our experiment by 1M, i.e., $N_{\max} = 1\text{M}$.

\addparagraphup\noindent{\bf Confidence calibration for sampling.}
For confidence calibration, we consider the following confidence loss $\mathcal{L}_{\mathtt{cnf}}$ to make the model produce confidence-calibrated outputs for data which are not relevant to the tasks the model learns:
\cutequationup
\begin{align*}
\mathcal{L}_{\mathtt{cnf}}(\theta, \phi; \mathcal{D}) &=
\frac1{\lvert \mathcal{D} \rvert \lvert \mathcal{T} \rvert} \sum_{x \in \mathcal{D}} \sum_{y \in \mathcal{T}}
[ -\log p(y | x ; \theta, \phi) ].
\end{align*}
\cutequationdown
During the 3-step learning, only the first step for training $\mathcal{C}_{t}$
has no reference model, so it should learn with the confidence loss.
For $\mathcal{C}_{t}$, $(x,y)$ is from OOD if $y \notin \mathcal{T}_{t}$.
Namely, by optimizing the confidence loss under the coreset of the previous tasks $\mathcal{D}_{t-1}^{\mathtt{cor}}$ and the external dataset $\mathcal{D}_{t}^{\mathtt{ext}}$, the model learns to produce a prediction with low confidence for OOD data, i.e., uniformly distributed probabilities over class labels.
Thus, $\mathcal{C}_{t}$ learns by optimizing the following:
\cutequationup
\begin{align}
\mathcal{L}_{\mathtt{cls}}(\theta^\mathcal{C}, \phi_{t}^\mathcal{C}; \mathcal{D}_{t})
&+ \mathcal{L}_{\mathtt{cnf}}(\theta^\mathcal{C}, \phi_{t}^\mathcal{C}; \mathcal{D}_{t-1}^{\mathtt{cor}} \cup \mathcal{D}_{t}^{\mathtt{ext}}).
\label{eq:tc_ex}
\end{align}
\cutequationdown
Note that the model $\mathcal{M}_{t}$ does not require an additional confidence calibration,
because
the previous model $\mathcal{P}_{t}$ is expected to be confidence-calibrated in the previous stage.
Therefore, the confidence-calibrated outputs of the reference models are distilled to the model $\mathcal{M}_{t}$.
The effect of confidence loss can be found in Table~\ref{tb:tc}.


\cutsectionup
\section{Related Work} \label{sec:related}
\cutsectiondown

\addparagraphup\noindent{\bf Continual lifelong learning.}
Many recent works have addressed catastrophic forgetting with different assumptions.
Broadly speaking, there are three different types of works \cite{van2018generative}:
one is class-incremental learning \cite{castro2018end, rebuffi2017icarl, wu2018incremental}, where the number of class labels keeps growing.
Another is task-incremental learning \cite{hou2018lifelong, li2016learning}, where the boundaries among tasks are assumed to be clear and the information about the task under test is given.\footnote{The main difference between class- and task-incremental learning is that the model has single- and multi-head output layer, respectively.}
The last can be seen as data-incremental learning, which is the case when the set of class labels or actions are the same for all tasks \cite{kirkpatrick2017overcoming, rusu2016progressive, schwarz2018progress}.

These works can be summarized as continual learning, and recent works on continual learning have studied two types of approaches to overcome catastrophic forgetting:
model-based and data-based.
Model-based approaches
\cite{aljundi2018memory, chaudhry2018riemannian, jung2018less, kirkpatrick2017overcoming, lee2017overcoming, lopez2017gradient, mallya2018piggyback, nguyen2018variational, ritter2018online, rusu2016progressive, schwarz2018progress, serra2018overcoming, yoon2017lifelong, zenke2017continual}
keep the knowledge of previous tasks by
penalizing the change of parameters crucial for previous tasks,
i.e., the updated parameters are constrained to be around the original values, and the update is scaled down by the importance of parameters on previous tasks.
However, since DNNs have many local optima, there would be better local optima for both the previous and new tasks, which cannot be found by model-based approaches.

On the other hand, data-based approaches \cite{castro2018end, hou2018lifelong, javed2018revisiting, li2016learning, rebuffi2017icarl} keep the knowledge of the previous tasks by knowledge distillation \cite{hinton2015distilling}, which minimizes the distance between the manifold of the latent space in the previous and new models.
In contrast to model-based approaches, they require to feed data to get features on the latent space.
Therefore, the amount of knowledge kept by knowledge distillation depends on the degree of similarity between the data distribution used to learn the previous tasks in the previous stages and the one used to distill the knowledge in the later stages.
To guarantee to have a certain amount of similar data, some prior works \cite{castro2018end, nguyen2018variational, rebuffi2017icarl} reserved a small amount of memory to keep a coreset,
and others \cite{lesort2018generative, rannen2017encoder, shin2017continual, van2018generative, wu2018incremental} trained a generative model and replay the generated data when training a new model.
Note that the model-based and data-based approaches are orthogonal in most cases, thus they can be combined for better performance \cite{kim2018keep}.

\addparagraphup\noindent{\bf Knowledge distillation in prior works.}
Our proposed method is a data-based approach, but it is different from prior works \cite{castro2018end, hou2018lifelong, li2016learning, rebuffi2017icarl}, because their model commonly learns with
the task-wise local distillation loss in Eq.~\eqref{eq:ld}.
We emphasize that local distillation only preserves the knowledge within each of the previous tasks, while global distillation does the knowledge over all tasks.


Similar to our 3-step learning, \citet{schwarz2018progress} and \citet{hou2018lifelong} utilized the idea of learning with two teachers.
However,
their strategy to keep the knowledge of the previous tasks is different:
\citet{schwarz2018progress} applied a model-based approach, and \citet{hou2018lifelong} distilled the task-wise knowledge for task-incremental learning.

On the other hand, \citet{castro2018end} had a similar fine-tuning, but they built a balanced dataset
by discarding most of the data of the current task and updated the whole networks.
However, such undersampling sacrifices the diversity of the frequent classes, which decreases the performance.
Oversampling may solve the issue, but it makes the training not scalable:
the size of the oversampled dataset increases proportional to the number of tasks learned so far.
Instead, we propose to apply data weighting.

\addparagraphup\noindent{\bf Scalability.}
Early works on continual learning were not scalable since they kept all previous models \cite{aljundi2017expert, kirkpatrick2017overcoming, li2016learning, rusu2016progressive, yoon2017lifelong}.
However, recent works considered the scalability by minimizing the amount of task-specific parameters \cite{rebuffi2017icarl, schwarz2018progress}.
In addition, data-based methods require to keep either a coreset or a generative model to replay previous tasks.
Our method is a data-based approach, but it does not suffer from the scalability issue
since we utilize an external dataset sampled from a large stream of unlabeled data.
We note that unlike coreset, our external dataset does not require a permanent memory;
it is discarded after learning.

\cutsectionup
\section{Experiments} \label{sec:exp}


\subsection{Experimental Setup}
\cutsubsectiondown

\addparagraphup\noindent{\bf Compared algorithms.}
To provide an upper bound of the performance, we compare an \emph{oracle} method, which learns by optimizing Eq.~\eqref{eq:gc} while storing all training data of previous tasks and replaying them during training.
Also, as a \emph{baseline}, we provide the performance of a model learned without knowledge distillation.
Among prior works, three state-of-the-art methods are compared:
\emph{learning without forgetting (LwF)}~\cite{li2016learning}, 
\emph{distillation and retrospection (DR)}~\cite{hou2018lifelong}, and
\emph{end-to-end incremental learning (E2E)}~\cite{castro2018end}.
For fair comparison, we adapt LwF and DR for class-incremental setting, which are originally evaluated in task-incremental learning setting:
specifically, we extend the range of the classification loss, i.e.,
we optimize Eq.~\eqref{eq:gc}+\eqref{eq:ld} and Eq.~\eqref{eq:gc}+\eqref{eq:ld}+\eqref{eq:gd_c} for replication of them.

We do not compare model-based methods, because data-based methods are known to outperform them in class-incremental learning \cite{lesort2018generative, van2018generative}, and they are orthogonal to data-based methods, such that they can potentially be combined with our approaches for better performance \cite{kim2018keep}.

\addparagraphup\noindent{\bf Datasets.}
We evaluate the compared methods on CIFAR-100 \cite{krizhevsky2009learning} and ImageNet ILSVRC 2012 \cite{deng2009imagenet}, where all images are downsampled to $32 \times 32$ \cite{chrabaszcz2017downsampled}.
For CIFAR-100, similar to prior works \cite{castro2018end, rebuffi2017icarl}, we shuffle the classes uniformly at random and split the classes to build a sequence of tasks.
For ImageNet, we first sample 500 images per 100 randomly chosen classes for each trial, and then split the classes.
To evaluate the compared methods under the environment with a large stream of unlabeled data, we take two large datasets: the TinyImages dataset \cite{torralba200880} with 80M images and the entire ImageNet 2011 dataset with 14M images.
The classes appeared in CIFAR-100 and ILSVRC 2012 are excluded to avoid any potential advantage from them.
At each stage, our sampling algorithm gets unlabeled data from them uniformly at random to form an external dataset, until the number of retrieved samples is 1M.

Following the prior works, we divide the classes into splits of 5, 10, and 20 classes, such that there are 20, 10, and 5 tasks, respectively.
For each task size, we evaluate the compared methods ten times with different class orders (different set of classes in the case of ImageNet) and report the mean and standard deviation of the performance.

\begin{table*}[t]
\cuttablecaptionup
\caption{Comparison of methods on CIFAR-100 and ImageNet.
We report the mean and standard deviation of ten trials for CIFAR-100 and nine trials for ImageNet with different random seeds in \%. $\uparrow$ ($\downarrow$) indicates that the higher (lower) number is the better.}
\cuttablecaptiondown
\label{tb:summary}
\setlength{\tabcolsep}{0.84mm}
\cuttableup
\centering\footnotesize
\begin{tabular}{c|c|c|c|c|c|c|c|c|c|c|c|c}
\toprule
Dataset & \multicolumn{6}{c|}{CIFAR-100} & \multicolumn{6}{c}{ImageNet} \cr
\hline
Task size & \multicolumn{2}{c|}{5} & \multicolumn{2}{c|}{10} & \multicolumn{2}{c|}{20} & \multicolumn{2}{c|}{5} & \multicolumn{2}{c|}{10} & \multicolumn{2}{c}{20} \cr
\hline
Metric & ACC ($\uparrow$) & FGT ($\downarrow$) & ACC ($\uparrow$) & FGT ($\downarrow$) & ACC ($\uparrow$) & FGT ($\downarrow$) & ACC ($\uparrow$) & FGT ($\downarrow$) & ACC ($\uparrow$) & FGT ($\downarrow$) & ACC ($\uparrow$) & FGT ($\downarrow$) \cr
\hline
Oracle
& 78.6 $\pm$ \scriptsize{0.9} & 3.3 $\pm$ \scriptsize{0.2} & 77.6 $\pm$ \scriptsize{0.8} & 3.1 $\pm$ \scriptsize{0.2} & 75.7 $\pm$ \scriptsize{0.7} & 2.8 $\pm$ \scriptsize{0.2} & 68.0 $\pm$ \scriptsize{1.7} & 3.3 $\pm$ \scriptsize{0.2} & 66.9 $\pm$ \scriptsize{1.6} & 3.1 $\pm$ \scriptsize{0.3} & 65.1 $\pm$ \scriptsize{1.2} & 2.7 $\pm$ \scriptsize{0.2} \cr
\hline
Baseline
& 57.4 $\pm$ \scriptsize{1.2} & 21.0 $\pm$ \scriptsize{0.5} & 56.8 $\pm$ \scriptsize{1.1} & 19.7 $\pm$ \scriptsize{0.4} & 56.0 $\pm$ \scriptsize{1.0} & 18.0 $\pm$ \scriptsize{0.3} & 44.2 $\pm$ \scriptsize{1.7} & 23.6 $\pm$ \scriptsize{0.4} & 44.1 $\pm$ \scriptsize{1.6} & 21.5 $\pm$ \scriptsize{0.5} & 44.7 $\pm$ \scriptsize{1.2} & 18.4 $\pm$ \scriptsize{0.5} \cr
LwF \citet{li2016learning}
& 58.4 $\pm$ \scriptsize{1.3} & 19.3 $\pm$ \scriptsize{0.5} & 59.5 $\pm$ \scriptsize{1.2} & 16.9 $\pm$ \scriptsize{0.4} & 60.0 $\pm$ \scriptsize{1.0} & 14.5 $\pm$ \scriptsize{0.4} & 45.6 $\pm$ \scriptsize{1.9} & 21.5 $\pm$ \scriptsize{0.4} & 47.3 $\pm$ \scriptsize{1.5} & 18.5 $\pm$ \scriptsize{0.5} & 48.6 $\pm$ \scriptsize{1.2} & 15.3 $\pm$ \scriptsize{0.6} \cr
DR \citet{hou2018lifelong}
& 59.1 $\pm$ \scriptsize{1.4} & 19.6 $\pm$ \scriptsize{0.5} & 60.8 $\pm$ \scriptsize{1.2} & 17.1 $\pm$ \scriptsize{0.4} & 61.8 $\pm$ \scriptsize{0.9} & 14.3 $\pm$ \scriptsize{0.4} & 46.5 $\pm$ \scriptsize{1.6} & 22.0 $\pm$ \scriptsize{0.5} & 48.7 $\pm$ \scriptsize{1.6} & 18.8 $\pm$ \scriptsize{0.5} & 50.7 $\pm$ \scriptsize{1.2} & 15.1 $\pm$ \scriptsize{0.5} \cr
E2E \citet{castro2018end}
& 60.2 $\pm$ \scriptsize{1.3} & 16.5 $\pm$ \scriptsize{0.5} & 62.6 $\pm$ \scriptsize{1.1} & 12.8 $\pm$ \scriptsize{0.4} & 65.1 $\pm$ \scriptsize{0.8} & 8.9 $\pm$ \scriptsize{0.2} & 47.7 $\pm$ \scriptsize{1.9} & 17.9 $\pm$ \scriptsize{0.4} & 50.8 $\pm$ \scriptsize{1.5} & 13.4 $\pm$ \scriptsize{0.4} & 53.9 $\pm$ \scriptsize{1.2} & 8.8 $\pm$ \scriptsize{0.3} \cr
\hline
GD (Ours)
& 62.1 $\pm$ \scriptsize{1.2} & 15.4 $\pm$ \scriptsize{0.4} & 65.0 $\pm$ \scriptsize{1.1} & 12.1 $\pm$ \scriptsize{0.3} & 67.1 $\pm$ \scriptsize{0.9} & 8.5 $\pm$ \scriptsize{0.3} & 50.0 $\pm$ \scriptsize{1.7} & 16.8 $\pm$ \scriptsize{0.4} & 53.7 $\pm$ \scriptsize{1.5} & 12.8 $\pm$ \scriptsize{0.5} & 56.5 $\pm$ \scriptsize{1.2} & 8.4 $\pm$ \scriptsize{0.4} \cr
+ \tt{ext}
& {\bf 66.3 $\pm$ \scriptsize{1.2}} & {\bf 9.8 $\pm$ \scriptsize{0.3}} & {\bf 68.1 $\pm$ \scriptsize{1.1}} & {\bf 7.7 $\pm$ \scriptsize{0.3}} & {\bf 68.9 $\pm$ \scriptsize{1.0}} & {\bf 5.5 $\pm$ \scriptsize{0.4}} & {\bf 55.2 $\pm$ \scriptsize{1.8}} & {\bf 9.6 $\pm$ \scriptsize{0.4}} & {\bf 57.7 $\pm$ \scriptsize{1.6}} & {\bf 7.4 $\pm$ \scriptsize{0.3}} & {\bf 58.7 $\pm$ \scriptsize{1.2}} & {\bf 5.4 $\pm$ \scriptsize{0.3}} \cr
\bottomrule
\end{tabular}
\cuttabledown
\end{table*}

\begin{figure*}[t]
\centering
\hspace*{\fill}
\subfigure[ACC improvement by learning with external data]
{\includegraphics[width=\twoawidth]{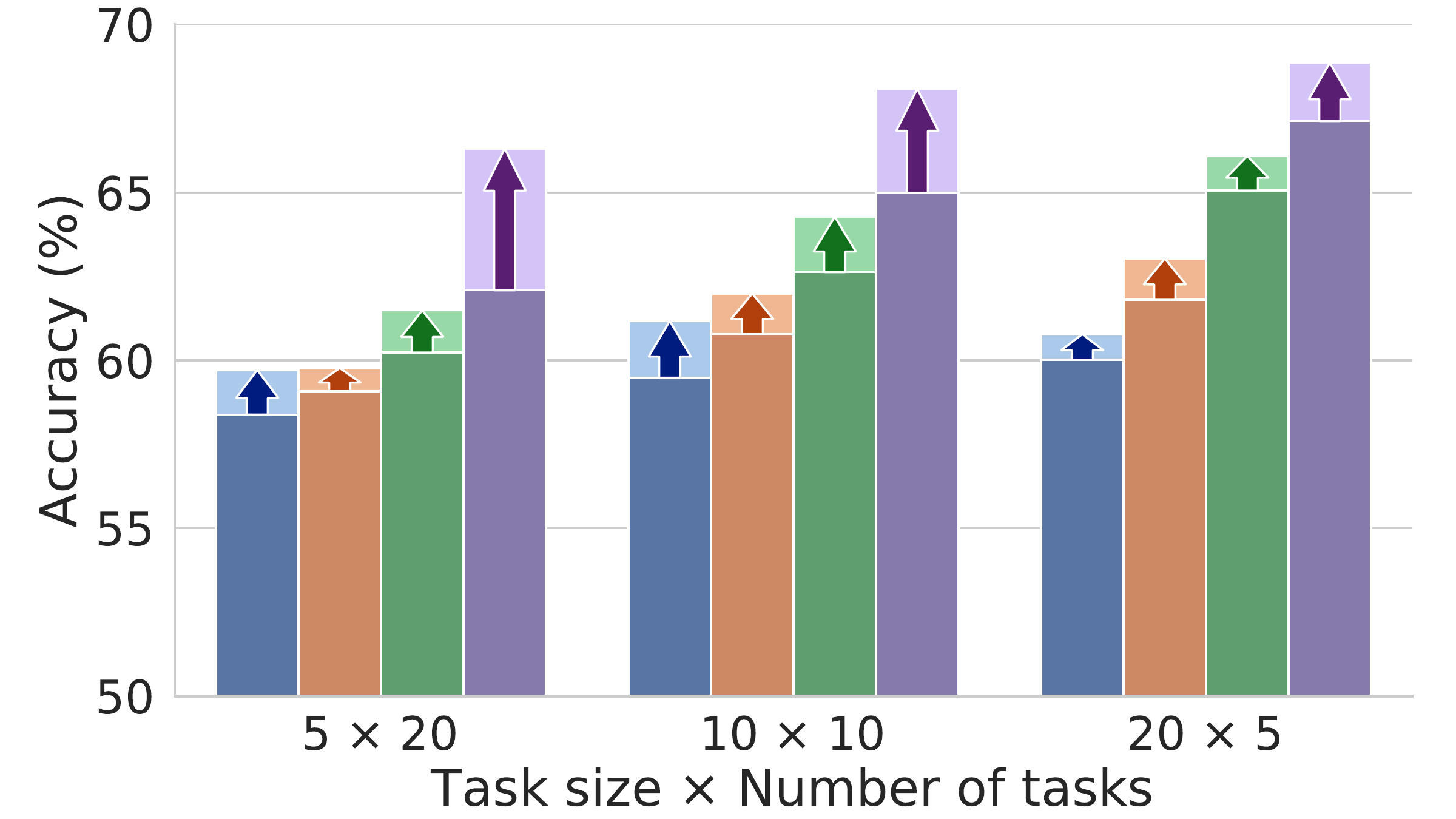}
\label{fig:acc}}
\hfill
\subfigure[FGT improvement by learning with external data \hspace{55pt}]
{\includegraphics[width=\twobwidth]{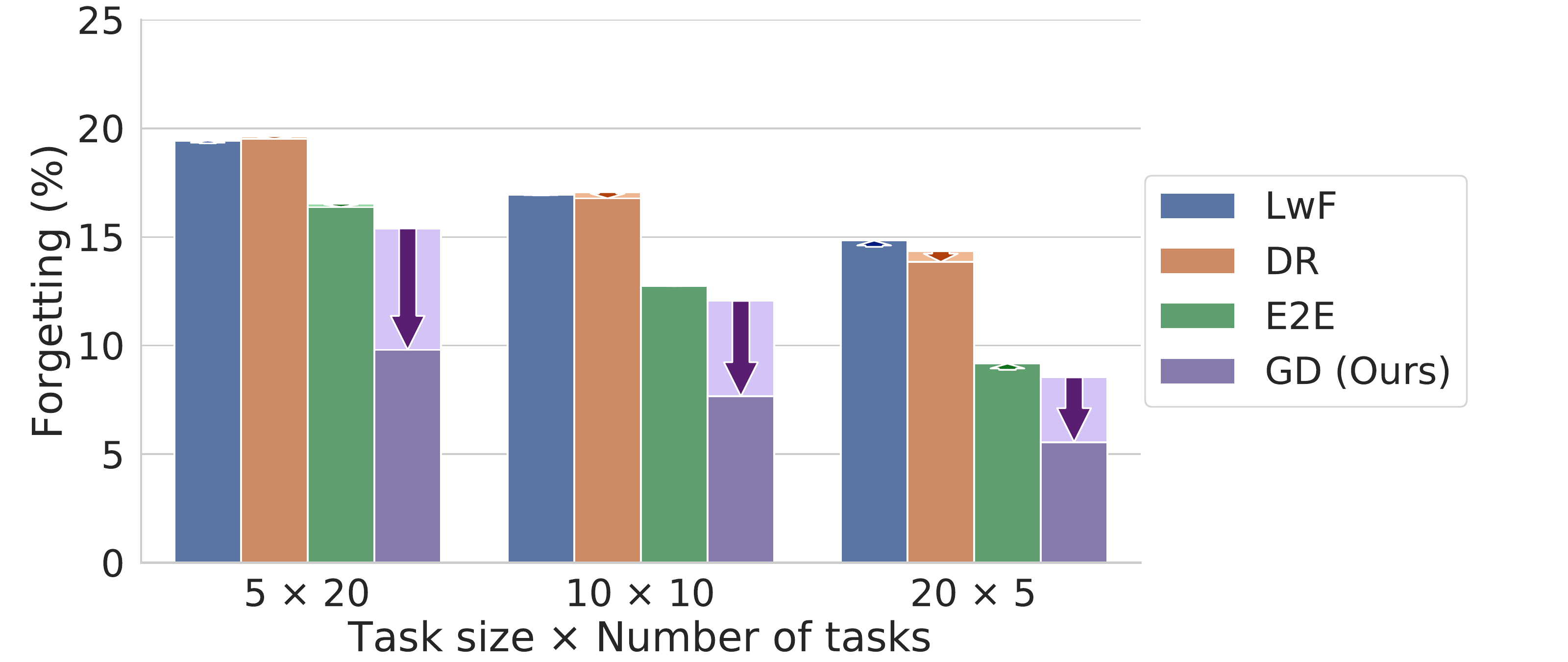}
\label{fig:fgt}} \\
\hspace*{\fill}
\subfigure[ACC with respect to the number of trained classes]
{\includegraphics[width=\twocwidth]{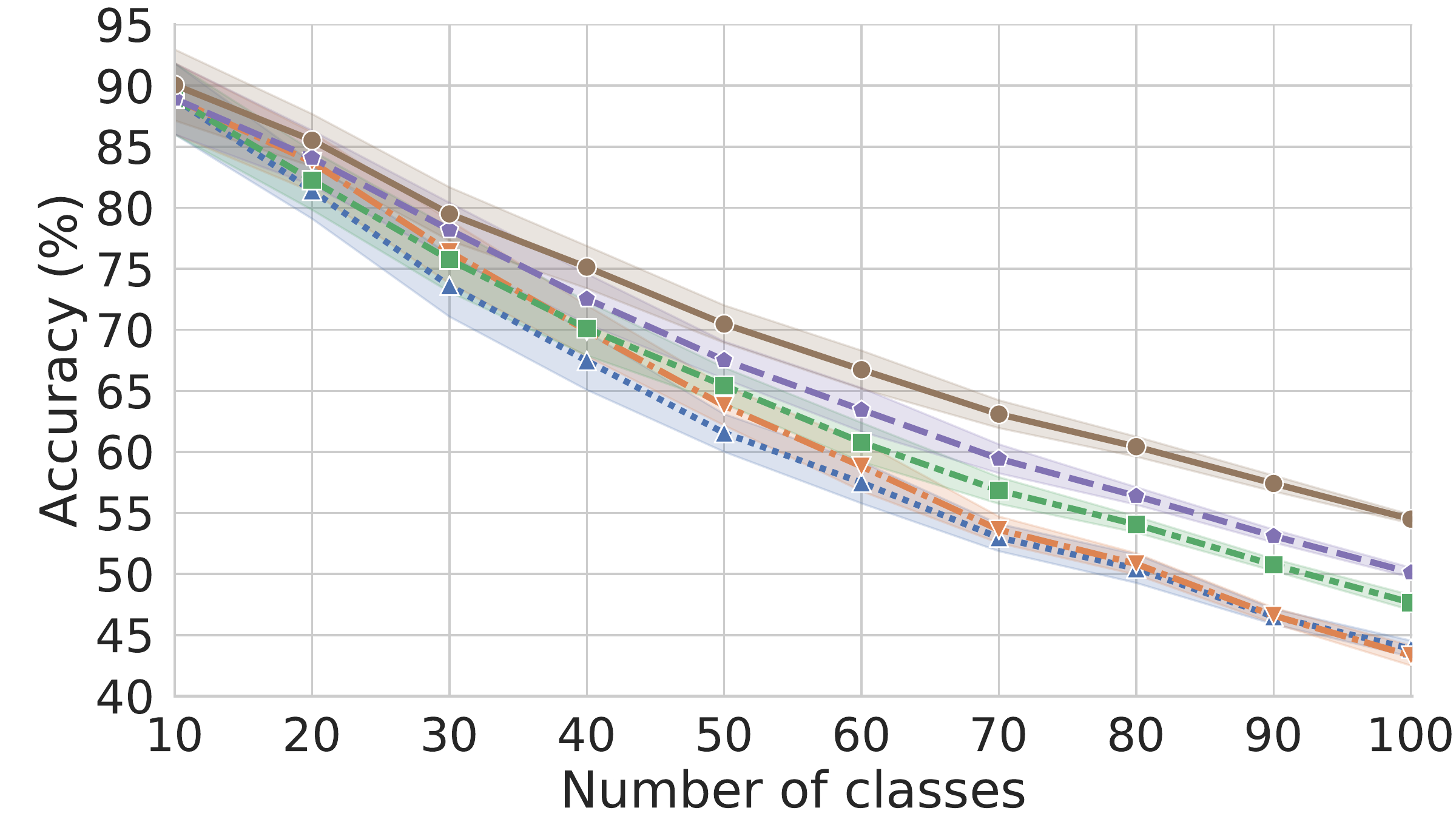}
\label{fig:acc_inc}}
\hfill
\subfigure[FGT with respect to the number of trained classes \hspace{55pt}]
{\includegraphics[width=\twodwidth]{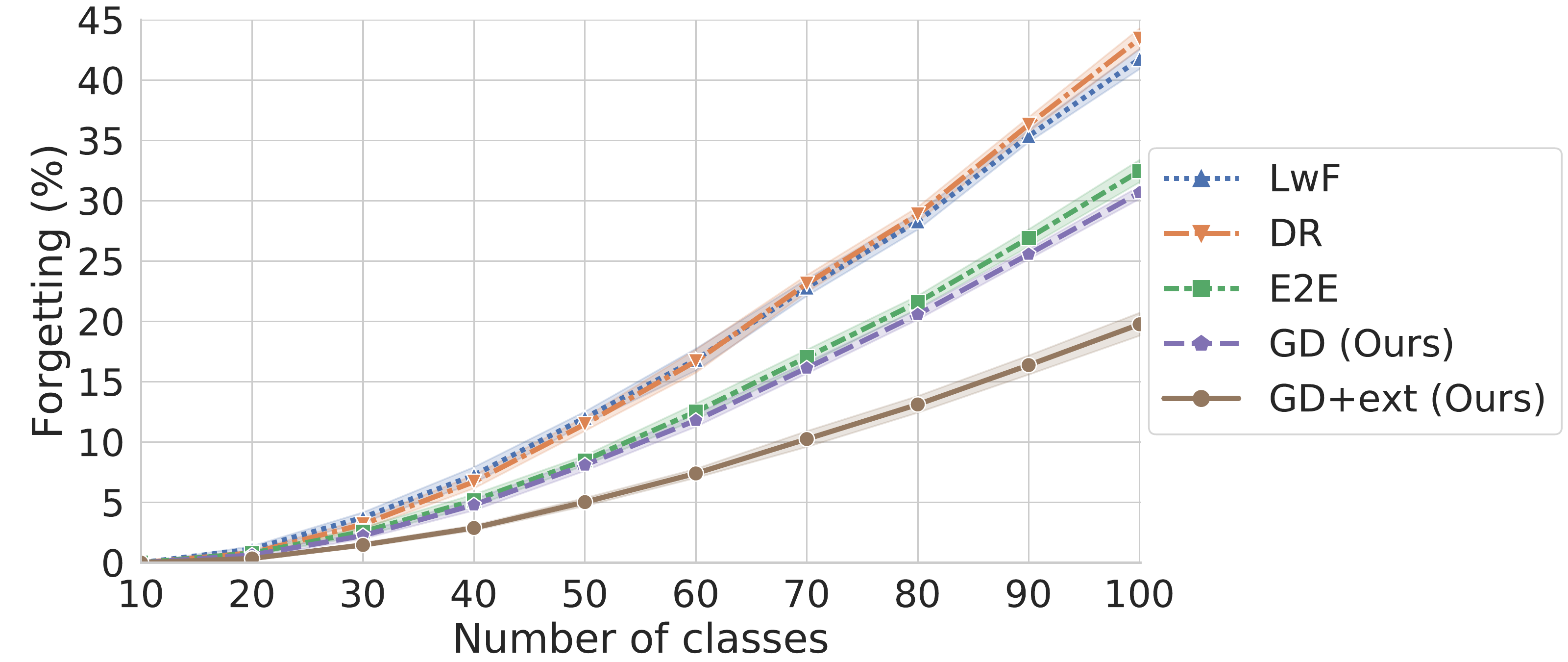}
\label{fig:fgt_inc}}
\cutcaptionup
\caption{
Experimental results on CIFAR-100.
(a,b) Arrows show the performance gain in the average incremental accuracy (ACC) and average forgetting (FGT) by learning with unlabeled data, respectively.
(c,d) Curves show ACC and FGT with respect to the number of trained classes when the task size is 10.
We report the average performance of ten trials.
}
\cutcaptiondown
\label{fig:exp}
\end{figure*}

\addparagraphup\noindent{\bf Evaluation metric.}
We report the performance of the compared methods in two metrics:
the average incremental accuracy (ACC) and the average forgetting (FGT).
For simplicity, we assume that the number of test data is the same over all classes.
For a test data from the $r$-th task $(x,y) \in \mathcal{D}_{r}^{\mathtt{test}}$, let $\hat{y}(x ; \mathcal{M}_{s})$ be the label predicted by the $s$-th model,
such that
\cutequationup
\begin{align*}
A_{r,s} = \frac1{\lvert \mathcal{D}_{r}^{\mathtt{test}} \rvert} \sum_{(x,y) \in \mathcal{D}_{r}^{\mathtt{test}}} \mathbb{I} (\hat{y}(x ; \mathcal{M}_{s}) = y)
\end{align*}
\cutequationdown
measures the accuracy of the $s$-th model at the $r$-th task, where $s \geq r$.
Note that prediction is done without task boundaries:
for example, at the $t$-th stage, the expected accuracy of random guess is $1 / \lvert \mathcal{T}_{1:t} \rvert$, not $1 / \lvert \mathcal{T}_{r} \rvert$.
At the $t$-th stage, ACC is defined as:
\cutequationup
\begin{align*}
\text{ACC} = 
\frac1{t-1} \sum_{s=2}^{t} \sum_{r=1}^{s} \frac{\lvert \mathcal{T}_{r} \rvert}{\lvert \mathcal{T}_{1:s} \rvert} A_{r,s}.
\end{align*}
\cutequationdown
Note that the performance of the first stage is not considered, as it is not class-incremental learning.
While ACC measures the overall performance directly, FGT measures the amount of catastrophic forgetting, by averaging the performance decay:
\cutequationup
\begin{align*}
\text{FGT} = 
\frac1{t-1} \sum_{s=2}^{t} \sum_{r=1}^{s-1} \frac{\lvert \mathcal{T}_{r} \rvert}{\lvert \mathcal{T}_{1:s} \rvert} (A_{r,r} - A_{r,s}),
\end{align*}
\cutequationdown
which is essentially the negative of the backward transfer \cite{lopez2017gradient}.
Note that smaller FGT is better, which implies that the model less-forgets about the previous tasks.


\addparagraphup\noindent{\bf Hyperparameters.}
The backbone of all compared models is
wide residual networks~\cite{zagoruyko2016wide} with 16 layers, a widen factor of 2 (WRN-16-2), and a dropout rate of 0.3.
Note that this has a comparable performance with ResNet-32~\cite{he2016deep}.
The last fully connected layer is considered to be a task-specific layer, and
whenever a task with new classes comes in, the layer is extended
to produce a prediction for the new classes.
The number of parameters in the task-specific layer is small compared to
those in shared layers (about 2\% in maximum in WRN-16-2).
All methods use the same size of coreset, which is 2000.
For scalability,
the size of the sampled external dataset is set to the size of the labeled dataset, i.e., $N_{D} = \lvert \mathcal{D}_{t}^{\mathtt{trn}} \rvert$ in Algorithm~\ref{alg:sample}.
For validation, one split of ImageNet is used, which is exclusive to the other nine trials.
The temperature for smoothing softmax probabilities \cite{hinton2015distilling} is set to 2 for distillation from $\mathcal{P}$ and $\mathcal{C}$,
and 1 for
$\mathcal{Q}$.
For more details, see \appendixword.

\cutsubsectionup
\subsection{Evaluation}
\cutsubsectiondown

\addparagraphup\noindent{\bf Comparison of methods.}
Table~\ref{tb:summary} and Figure~\ref{fig:exp} compare our proposed methods with the state-of-the-art methods.
First, even when unlabeled data are not accessible,
our method outperforms the state-of-the-art methods,
which shows the effectiveness of the proposed 3-step learning scheme.
Specifically, in addition to the difference in the loss function,
DR does not have balanced fine-tuning, E2E lacks the teacher for the current task $\mathcal{C}_{t}$ and fine-tunes the whole networks with a small dataset, and LwF has neither $\mathcal{C}_{t}$ nor fine-tuning.
Compared to E2E, which is the best state-of-the-art method, our method improves ACC by 4.8\% and FGT by 6.0\% on ImageNet with a task size of 5.

On the other hand,
as shown in Figure~\ref{fig:acc}--\ref{fig:fgt},
learning with an unlabeled external dataset improves the performance of compared methods consistently, but the improvement is more significant in GD.
For example, in the case of ImageNet with a task size of 5, by learning with the external dataset, E2E improves ACC by 3.2\%, while GD does by 10.5\%.
Also, the relative performance gain in terms of FGT is more significant:
E2E forgets 1.1\% less while GD does 43.1\%.
Overall, with our proposed learning scheme and knowledge distillation with the external dataset, GD improves its ACC by 15.8\% and FGT by 46.5\% over E2E.



\begin{table}[t]
\cuthalftablecaptionup
\caption{
Comparison of models learned with different reference models on CIFAR-100 when the task size is 10.
``$\mathcal{P}$,'' ``$\mathcal{C}$,'' and ``$\mathcal{Q}$'' stand for the previous model, the teacher for the current task, and their ensemble model, respectively.
}
\cuthalftablecaptiondown
\label{tb:loss}
\cuthalftableup
\centering
\begin{tabular}{c|c|c|c|c}
\toprule
$\mathcal{P}$ & $\mathcal{C}$ & $\mathcal{Q}$ & ACC ($\uparrow$) & FGT ($\downarrow$) \cr
\hline
\checkmark & &
& 62.9 $\pm$ \scriptsize{1.2} & 14.7 $\pm$ \scriptsize{0.4} \cr
\checkmark & \checkmark &
& 67.0 $\pm$ \scriptsize{0.9} & 10.7 $\pm$ \scriptsize{0.3} \cr
& & \checkmark
& 65.7 $\pm$ \scriptsize{0.9} & 11.2 $\pm$ \scriptsize{0.2} \cr
\checkmark & \checkmark & \checkmark
& {\bf 68.1 $\pm$ \scriptsize{1.1}} & {\bf 7.7 $\pm$ \scriptsize{0.3}} \cr
\bottomrule
\end{tabular}
\cuthalftabledown
\end{table}

\begin{table}[t]
\cuthalftablecaptionup
\caption{
Comparison of models learned
with a different teacher
for the current task $\mathcal{C}$
on CIFAR-100 when the task size is 10.
For ``{\tt cls},'' $\mathcal{C}$ is not trained but
the model learns by optimizing the learning objective of $\mathcal{C}$ directly.
The model learns with the proposed 3-step learning for ``{\tt dst}.''
The confidence loss is added to the learning objective for $\mathcal{C}$ for ``{\tt cnf}.''
We do not utilize $\mathcal{Q}$ for this experiment, because ``{\tt cls}'' has no explicit $\mathcal{C}$.
}
\cuthalftablecaptiondown
\label{tb:tc}
\cuthalftableup
\centering
\begin{tabular}{c|c|c|c}
\toprule
$\mathcal{C}$ & Confidence & ACC ($\uparrow$) & FGT ($\downarrow$) \cr
\hline
\xmark    &           & 62.9 $\pm$ \scriptsize{1.2} & 14.7 $\pm$ \scriptsize{0.4} \cr
{\tt cls} &           & 62.9 $\pm$ \scriptsize{1.3} & 14.5 $\pm$ \scriptsize{0.5} \cr
{\tt cls} & {\tt cnf} & 65.3 $\pm$ \scriptsize{1.0} & 11.7 $\pm$ \scriptsize{0.3} \cr
{\tt dst} &           & 66.2 $\pm$ \scriptsize{1.0} & 11.2 $\pm$ \scriptsize{0.3} \cr
{\tt dst} & {\tt cnf} & {\bf 67.0 $\pm$ \scriptsize{0.9}} & {\bf 10.7 $\pm$ \scriptsize{0.3}} \cr
\bottomrule
\end{tabular}
\cuthalftabledown
\end{table}

\begin{table}[t]
\cuthalftablecaptionup
\caption{
Comparison of different balanced learning strategies on CIFAR-100 when the task size is 10.
``DW,'' ``FT-DSet,'' and ``FT-DW'' stand for training with data weighting in Eq.~\eqref{eq:dw} for the entire training, fine-tuning with a training dataset balanced by removing data of the current task,
and fine-tuning with data weighting, respectively.
}
\cuthalftablecaptiondown
\label{tb:balance}
\cuthalftableup
\centering
\begin{tabular}{c|c|c}
\toprule
Balancing & ACC ($\uparrow$) & FGT ($\downarrow$) \cr
\hline
\xmark  & 67.1 $\pm$ \scriptsize{0.9} & 11.5 $\pm$ \scriptsize{0.3} \cr
DW      & 67.9 $\pm$ \scriptsize{0.9} & 9.6 $\pm$ \scriptsize{0.2} \cr
FT-DSet & 67.2 $\pm$ \scriptsize{1.1} & 8.4 $\pm$ \scriptsize{0.2} \cr
FT-DW   & {\bf 68.1 $\pm$ \scriptsize{1.1}} & {\bf 7.7 $\pm$ \scriptsize{0.3}} \cr
\bottomrule
\end{tabular}
\cuthalftabledown
\end{table}

\addparagraphup\noindent{\bf Effect of the reference models.}
Table~\ref{tb:loss} shows an ablation study with different set of reference models.
As discussed in Section~\ref{sec:gd}, because the previous model $\mathcal{P}$ does not know about the current task, the compensation by introducing $\mathcal{C}$ improves the overall performance.
On the other hand, $\mathcal{Q}$ does not show better ACC than the combination of $\mathcal{P}$ and $\mathcal{C}$.
This would be because, when building the output of $\mathcal{Q}$, the ensemble of the output of $\mathcal{P}$ and $\mathcal{C}$ is made with an assumption, which would not always be true.
However, the knowledge from $\mathcal{Q}$ is useful, such that the combination of all three reference models shows the best performance.

\addparagraphup\noindent{\bf Effect of the teacher for the current task $\mathcal{C}$.}
Table~\ref{tb:tc} compares the models learned
with a different teacher
for the current task $\mathcal{C}_{t}$.
In addition to the baseline without $\mathcal{C}_{t}$, we also 
compare
the model directly optimizes the learning objective of $\mathcal{C}_{t}$ in Eq.~\eqref{eq:tc} or \eqref{eq:tc_ex},
i.e., the model learns with hard labels rather than soft labels when optimizing that loss.
Note that introducing a separate model $\mathcal{C}$ for distillation is beneficial, because $\mathcal{C}$ learns better knowledge about the current task without interference from other classification tasks.
Learning by optimizing the confidence loss improves the performance,
because
the confidence-calibrated model samples better
external data as discussed in Section~\ref{sec:sample}.

\addparagraphup\noindent{\bf Effect of balanced fine-tuning.}
Table~\ref{tb:balance} shows the effect of balanced learning.
First, balanced learning strategies improve FGT in general.
If fine-tuning in 3-step learning is skipped but data weighting in Eq.~\eqref{eq:dw} is applied in the main training (DW), the model shows higher FGT than having balanced fine-tuning on task-specific parameters (FT-DW),
as discussed in Section~\ref{sec:gd}.
Note that data weighting (FT-DW) is better than removing the data of the current task to construct a small balanced dataset (FT-DSet)
proposed in \citet{castro2018end}, because all training data are useful.

\begin{table}[t]
\cuthalftablecaptionup
\caption{
Comparison of different external data sampling strategies on CIFAR-100 when the task size is 10.
``Prev'' and ``OOD'' columns describe
the sampling method for
data of previous tasks and out-of-distribution data, where
``Pred'' and ``Random'' stand for
sampling based on the prediction of the previous model $\mathcal{P}$ and random sampling, respectively.
In particular, for when sampling OOD by ``Pred,'' we sample data minimizing the confidence loss $\mathcal{L}_{\mathtt{cnf}}$.
When only Prev or OOD is sampled,
the number of sampled data is matched for fair comparison.
}
\cuthalftablecaptiondown
\label{tb:sample}
\cuthalftableup
\centering
\begin{tabular}{c|c|c|c}
\toprule
Prev & OOD & ACC ($\uparrow$) & FGT ($\downarrow$) \cr
\hline
\xmark & \xmark & 65.0 $\pm$ \scriptsize{1.1} & 12.1 $\pm$ \scriptsize{0.3} \cr
\xmark & Random & 67.6 $\pm$ \scriptsize{0.9} & 9.0 $\pm$ \scriptsize{0.3} \cr
Pred   & \xmark & 66.0 $\pm$ \scriptsize{1.2} & 7.8 $\pm$ \scriptsize{0.3} \cr
Pred   & Pred   & 65.7 $\pm$ \scriptsize{1.1} & 10.2 $\pm$ \scriptsize{0.2} \cr
Pred   & Random & {\bf 68.1 $\pm$ \scriptsize{1.1}} & {\bf 7.7 $\pm$ \scriptsize{0.3}} \cr
\bottomrule
\end{tabular}
\cuthalftabledown
\vspace*{-4pt}
\end{table}


\addparagraphup\noindent{\bf Effect of external data sampling.}
Table~\ref{tb:sample} compares different external data sampling strategies.
Unlabeled data are beneficial in all cases, but the performance gain is different over sampling strategies.
First, observe that randomly sampled data are useful, because their predictive distribution would be diverse such that it helps to
learn the diverse knowledge of the reference models, which makes the model confidence-calibrated.
However, while the random sampling strategy has higher ACC than sampling based on the prediction of the previous model $\mathcal{P}$, it also shows high FGT.
This implies that the unlabeled data sampled based on the prediction of $\mathcal{P}$ prevents the model from catastrophic forgetting more.
As discussed in Section~\ref{sec:sample}, our proposed sampling strategy, the combination of the above two strategies shows the best performance.
Finally,
sampling OOD data based on the prediction of $\mathcal{P}$ is not beneficial, because ``data most likely to be from OOD'' would not be useful.
OOD data sampled based on the prediction of $\mathcal{P}$ have almost uniform predictive distribution, which would be locally distributed.
However, the concept of OOD is a kind of complement set of the data distribution the model learns.
Thus, to learn to discriminate OOD well in our case, the model should learn with data widely distributed outside of the data distribution of the previous tasks.


\cutsectionup
\section{Conclusion} \label{sec:conclusion}
\cutsectiondown


We propose to leverage a large stream of unlabeled data in the wild for class-incremental learning.
The proposed global distillation aims to keep
the knowledge of the reference models without task boundaries, leading better knowledge distillation.
Our 3-step learning scheme effectively leverages the external dataset sampled by the confidence-based sampling strategy from the stream of unlabeled data.


\cutsectionup
\section*{Acknowledgements}
\cutsectiondown


This work was supported in part by
Kwanjeong Educational Foundation Scholarship,
NSF CAREER IIS-1453651, and
Sloan Research Fellowship.
We also thank Lajanugen Logeswaran, Jongwook Choi, Yijie Guo, Wilka Carvalho, and Yunseok Jang for helpful discussions.

{\small
\bibliographystyle{ieee_fullname}
\bibliography{ref}
}

\ifdefined\printappendix
\onecolumn
\clearpage
\part*{Appendix}
\appendix
\numberwithin{table}{section}
\numberwithin{figure}{section}
\numberwithin{equation}{section}

\section{Illustration of Global Distillation}

\begin{figure*}[ht]
\centering
\includegraphics[width=\aonewidth]{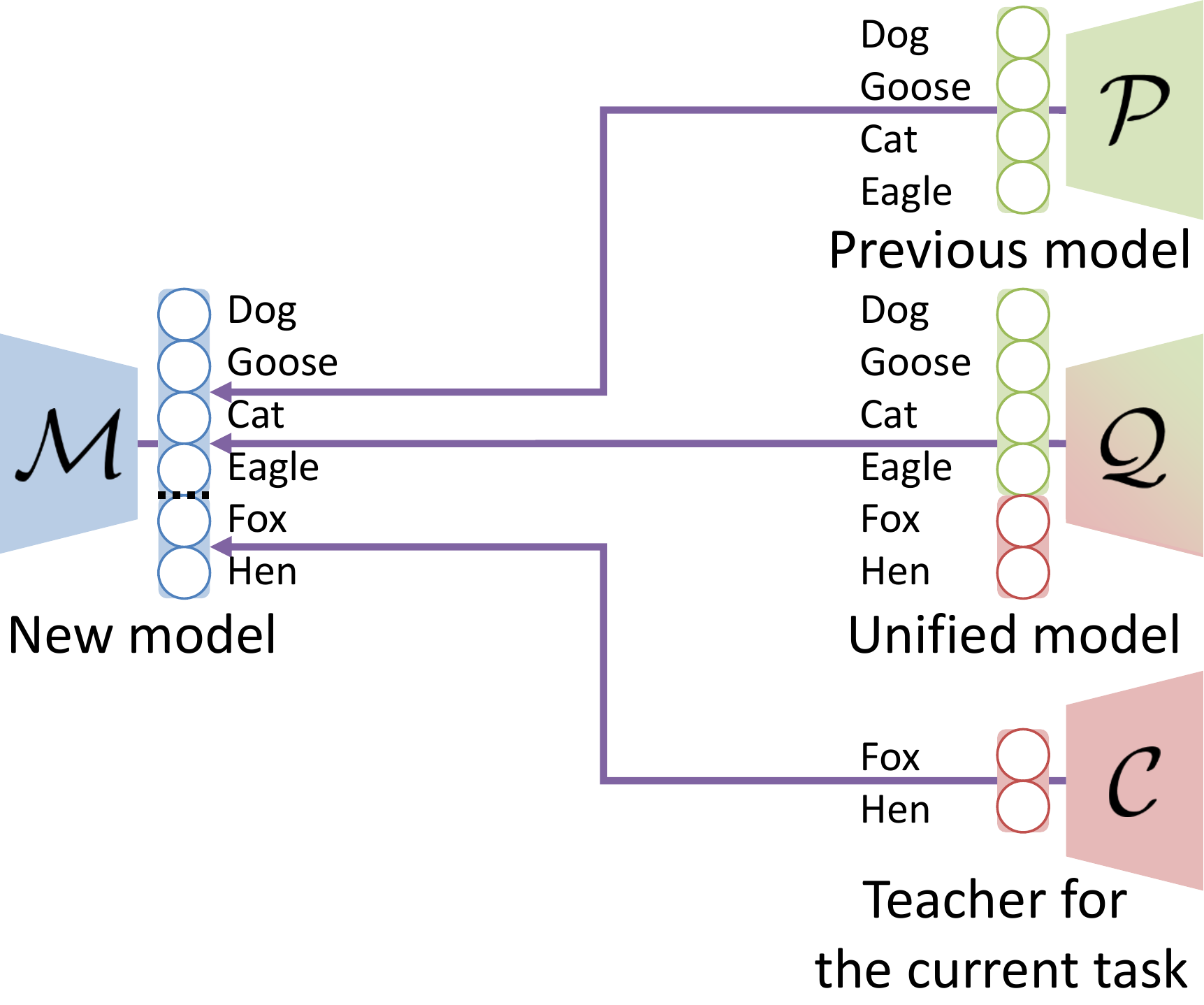}
\cuthalfcaptionup
\caption{
An illustration of how a model $\mathcal{M}$ learns with global distillation (GD).
For GD, three reference models are used:
$\mathcal{P}$ is the previous model,
$\mathcal{C}$ is the teacher for the current task, and
$\mathcal{Q}$ is an ensemble of them.
}
\cuthalfcaptiondown
\label{fig:pcq}
\end{figure*}

\section{Details on Experimental Setup}

\addparagraphup\noindent{\bf Hyperparameters.}
We use mini-batch training with a batch size of 128 over 200 epochs for each training to ensure convergence.
The initial learning rate is 0.1 and decays by 0.1 after 120, 160, 180 epochs when there is no fine-tuning.
When fine-tuning is applied, the model is first trained over 180 epochs where the learning rate decays after 120, 160, 170 epochs, and then fine-tuned over 20 epochs, where the learning rate starts at 0.01 and decays by 0.1 after 10, 15 epochs.
We note that 20 epochs are enough for convergence even when fine-tuning the whole networks for some methods.
We update the model parameters by stochastic gradient decent with a momentum 0.9 and an L2 weight decay of 0.0005.
The size of the coreset is set to 2000.
Due to the scalability issue, the size of the sampled external dataset is set to the size of the labeled dataset.
The ratio of OOD data in sampling is determined by validation on a split of ImageNet, which is 0.7.
For all experiments, the temperature for smoothing softmax probabilities is set to 2 for distillation from $\mathcal{P}$ and $\mathcal{C}$, and 1 for distillation from $\mathcal{Q}$.
To be more specific about the way to scale probabilities,
let $z = \{ z_{y} | y \in \mathcal{T} \} = \mathcal{M}(x; \theta, \phi)$ be the set of outputs (or logits).
Then, with a temperature $\gamma$, the probabilities are computed as follows:
\begin{align*}
p(y=k|x, \theta, \phi) = \frac{\exp(z_{k} / \gamma)}{\sum_{y' \in \mathcal{T}} \exp(z_{y'} / \gamma)}.
\end{align*}

\addparagraphup\noindent{\bf Scalability of methods.}
We note that all compared methods are scalable and they are compared in a fair condition.
We do not compare generative replay methods with ours, because the coreset approach is known to outperform them in class-incremental learning in a scalable setting:
in particular, it has been reported that continual learning for a generative model is a challenging problem on datasets of natural images like CIFAR-100 \cite{lesort2018generative, wu2018incremental}.

\clearpage

\section{More Experimental Results}

\subsection{More Ablation Studies}

\begin{figure*}[ht]
\centering
\hspace*{\fill}
\subfigure[CIFAR-100 ACC]
{\includegraphics[width=\conewidth]{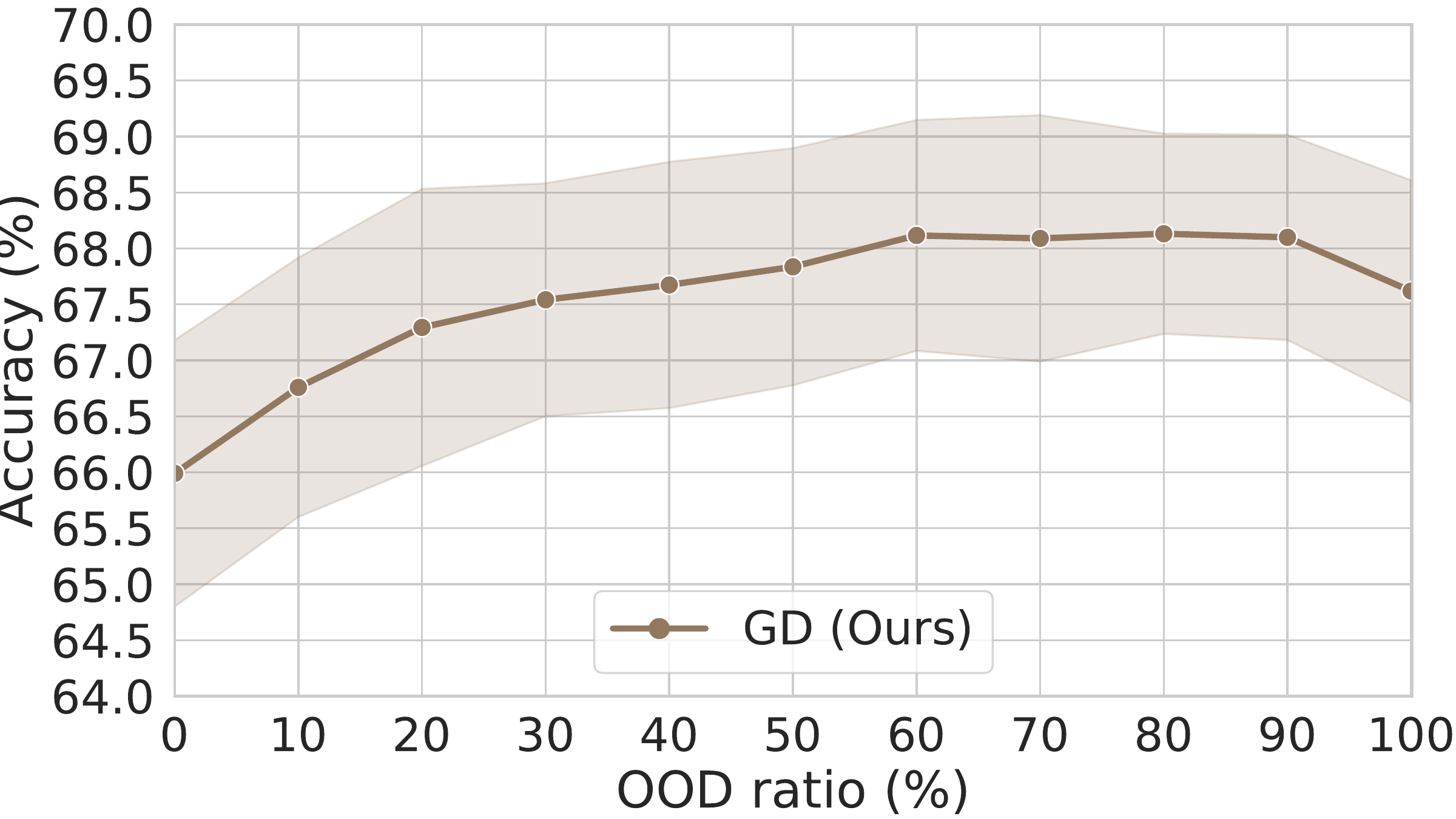}
\label{fig:acc_cifar100_ood_ratio}}
\hfill
\subfigure[CIFAR-100 FGT]
{\includegraphics[width=\conewidth]{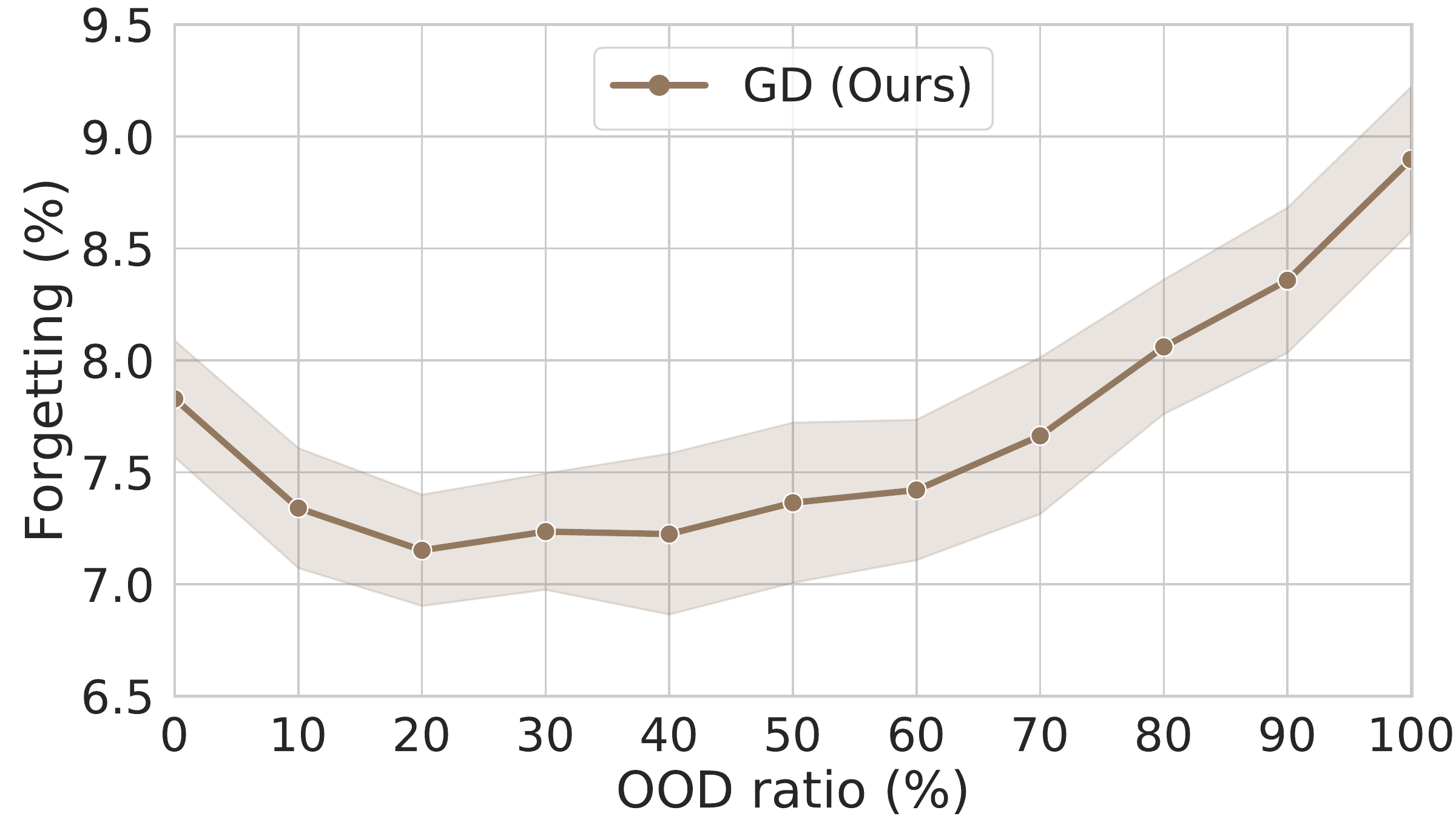}
\label{fig:fgt_cifar100_ood_ratio}}
\hspace*{\fill} \\
\hspace*{\fill}
\subfigure[ImageNet ACC]
{\includegraphics[width=\conewidth]{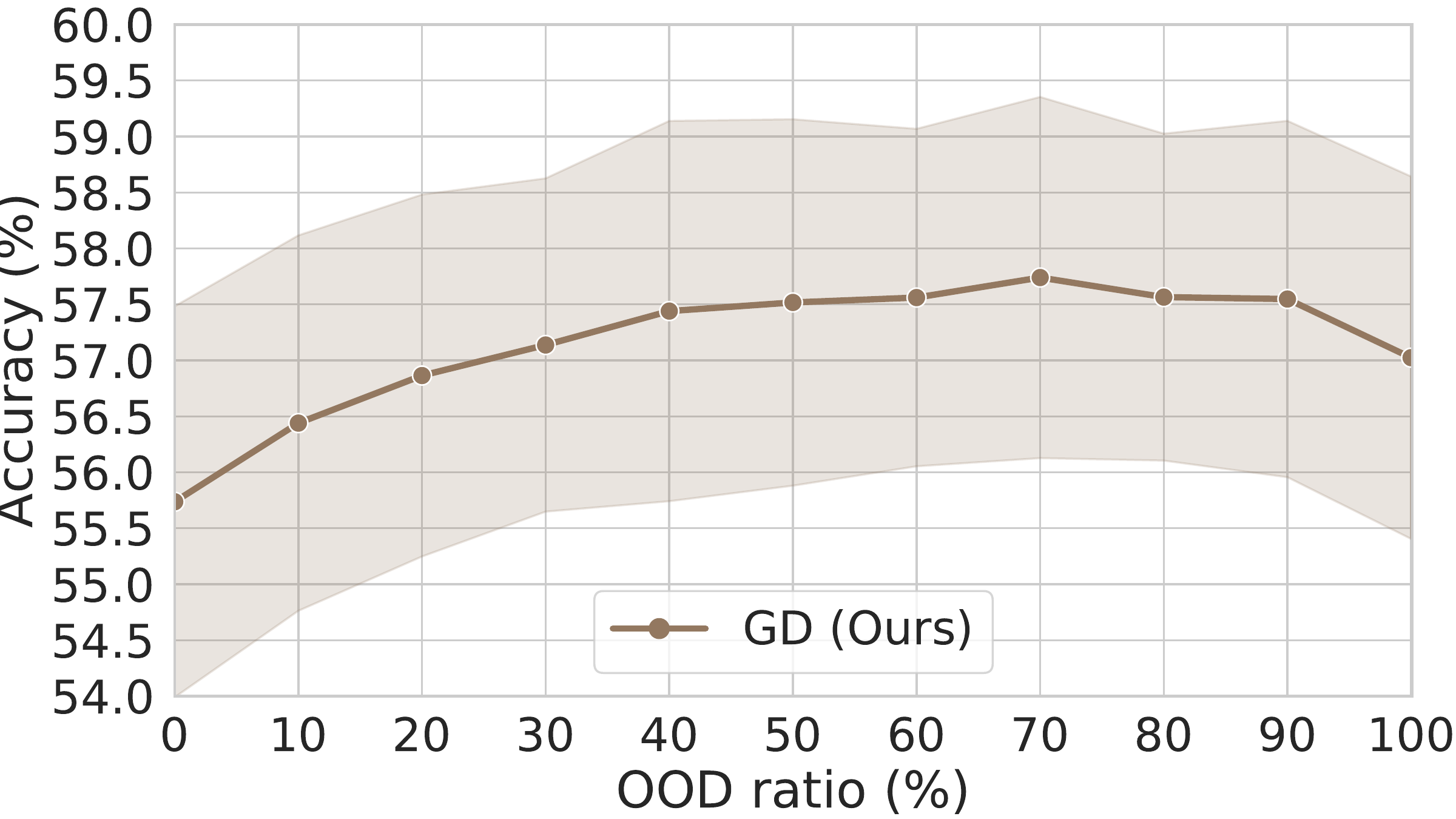}
\label{fig:acc_imagenet_ood_ratio}}
\hfill
\subfigure[ImageNet FGT]
{\includegraphics[width=\conewidth]{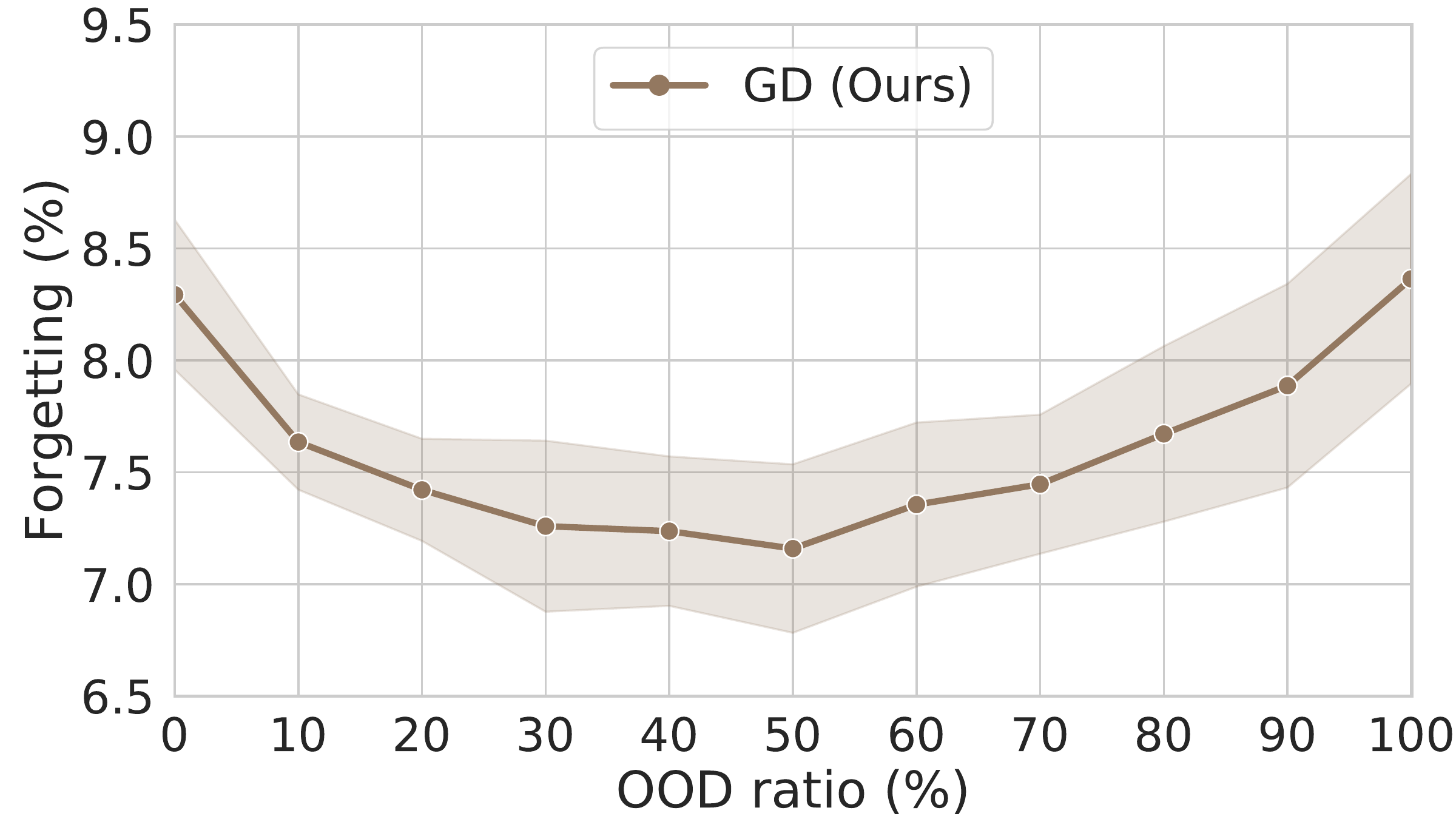}
\label{fig:fgt_imagenet_ood_ratio}}
\hspace*{\fill}
\cutcaptionup
\caption{
Experimental results on CIFAR-100 and ImageNet when the task size is 10.
We report ACC and FGT with respect to the OOD ratio averaged over ten trials for CIFAR-100 and nine trials for ImageNet.
}
\cutcaptiondown
\label{fig:exp_ood_ratio}
\end{figure*}

\addparagraphup\noindent{\bf Effect of the OOD ratio.}
We investigate the effect of the ratio between the sampled data likely to be in the previous tasks and OOD data.
As shown in Figure~\ref{fig:exp_ood_ratio}, the optimal OOD ratio varies over datasets, but it is higher than 0.5:
specifically, the best ACC is achieved when the OOD ratio is 0.8 on CIFAR-100, and 0.7 on ImageNet.
On the other hand, the optimal OOD ratio for FGT is different:
specifically, the best FGT is achieved when the OOD ratio is 0.2 on CIFAR-100, and 0.5 on ImageNet.

\begin{figure*}[ht]
\centering
\hspace*{\fill}
\subfigure[ImageNet ACC]
{\includegraphics[width=\ctwowidth]{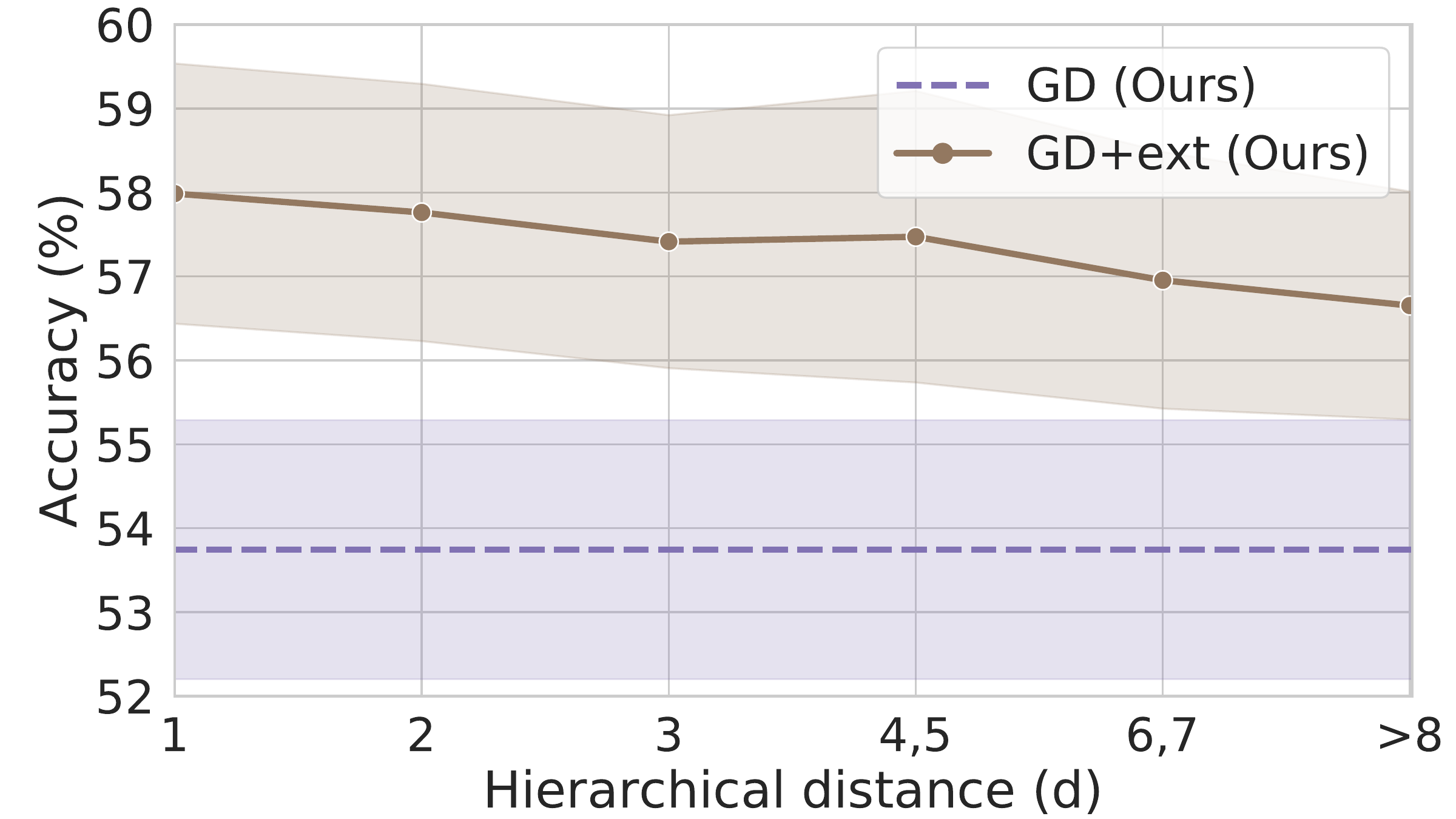}
\label{fig:acc_ex_dist_1d_imagenet}}
\hfill
\subfigure[ImageNet FGT]
{\includegraphics[width=\ctwowidth]{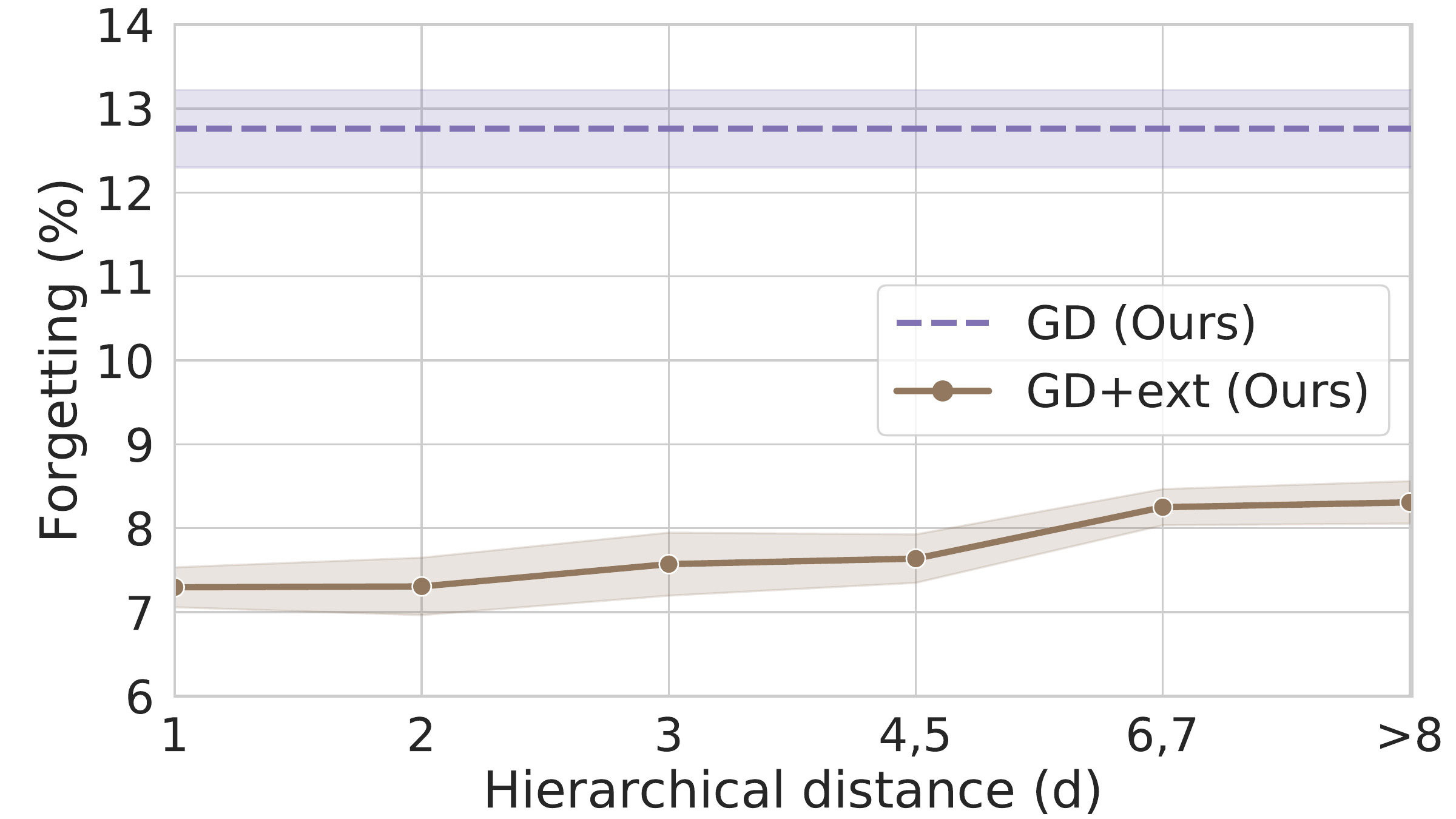}
\label{fig:fgt_ex_dist_1d_imagenet}}
\hspace*{\fill}
\label{fig:ex_dist_1d_imagenet}
\cutcaptionup
\caption{
Experimental results on ImageNet when the task size is 10.
We report ACC and FGT with respect to the hierarchical distance between the training dataset and unlabeled data stream averaged over nine trials.
}
\cutcaptiondown
\label{fig:exp_ex_dist_imagenet}
\end{figure*}

\addparagraphup\noindent{\bf Effect of the correlation between the training data and unlabeled external data.}
So far, we do not assume any correlation between training data and external data.
However, in this experiment, we control the correlation between them based on the hypernym-hyponym relationship between ImageNet class labels.
Specifically, we first compute the hierarchical distance (the length of the shortest path between classes in hierarchy) between 1k classes in ImageNet ILSVRC 2012 training dataset and the other 21k classes in the entire ImageNet 2011 dataset.
Note that the hierarchical distance can be thought as the semantic difference between classes.
Then, we divide the 21k classes based on the hierarchical distance, such that each split has at least 1M images for simulating an unlabeled data stream.
As shown in Figure~\ref{fig:exp_ex_dist_imagenet}, the performance is proportional to the semantic similarity, which is inversely proportional the hierarchical distance.
However, even in the worst case, unlabeled data are beneficial.

\clearpage

\subsection{More Results}

\begin{table*}[ht]
\cuttablecaptionup
\caption{Comparison of methods on CIFAR-100 and ImageNet.
We report the mean and standard deviation of ten trials for CIFAR-100 and nine trials for ImageNet with different random seeds in \%. $\uparrow$ ($\downarrow$) indicates that the higher (lower) number is the better.}
\cuttablecaptiondown
\label{tb:summary_full}
\setlength{\tabcolsep}{0.84mm}
\cuttableup
\centering\footnotesize
\begin{tabular}{c|c|c|c|c|c|c|c|c|c|c|c|c}
\toprule
Dataset & \multicolumn{6}{c|}{CIFAR-100} & \multicolumn{6}{c}{ImageNet} \cr
\hline
Task size & \multicolumn{2}{c|}{5} & \multicolumn{2}{c|}{10} & \multicolumn{2}{c|}{20} & \multicolumn{2}{c|}{5} & \multicolumn{2}{c|}{10} & \multicolumn{2}{c}{20} \cr
\hline
Metric & ACC ($\uparrow$) & FGT ($\downarrow$) & ACC ($\uparrow$) & FGT ($\downarrow$) & ACC ($\uparrow$) & FGT ($\downarrow$) & ACC ($\uparrow$) & FGT ($\downarrow$) & ACC ($\uparrow$) & FGT ($\downarrow$) & ACC ($\uparrow$) & FGT ($\downarrow$) \cr
\hline
Oracle
& 78.6 $\pm$ \scriptsize{0.9} & 3.3 $\pm$ \scriptsize{0.2} & 77.6 $\pm$ \scriptsize{0.8} & 3.1 $\pm$ \scriptsize{0.2} & 75.7 $\pm$ \scriptsize{0.7} & 2.8 $\pm$ \scriptsize{0.2} & 68.0 $\pm$ \scriptsize{1.7} & 3.3 $\pm$ \scriptsize{0.2} & 66.9 $\pm$ \scriptsize{1.6} & 3.1 $\pm$ \scriptsize{0.3} & 65.1 $\pm$ \scriptsize{1.2} & 2.7 $\pm$ \scriptsize{0.2} \cr
\hline
\multicolumn{3}{c}{Without an external dataset} \cr
Baseline
& 57.4 $\pm$ \scriptsize{1.2} & 21.0 $\pm$ \scriptsize{0.5} & 56.8 $\pm$ \scriptsize{1.1} & 19.7 $\pm$ \scriptsize{0.4} & 56.0 $\pm$ \scriptsize{1.0} & 18.0 $\pm$ \scriptsize{0.3} & 44.2 $\pm$ \scriptsize{1.7} & 23.6 $\pm$ \scriptsize{0.4} & 44.1 $\pm$ \scriptsize{1.6} & 21.5 $\pm$ \scriptsize{0.5} & 44.7 $\pm$ \scriptsize{1.2} & 18.4 $\pm$ \scriptsize{0.5} \cr
LwF \citet{li2016learning}
& 58.4 $\pm$ \scriptsize{1.3} & 19.3 $\pm$ \scriptsize{0.5} & 59.5 $\pm$ \scriptsize{1.2} & 16.9 $\pm$ \scriptsize{0.4} & 60.0 $\pm$ \scriptsize{1.0} & 14.5 $\pm$ \scriptsize{0.4} & 45.6 $\pm$ \scriptsize{1.9} & 21.5 $\pm$ \scriptsize{0.4} & 47.3 $\pm$ \scriptsize{1.5} & 18.5 $\pm$ \scriptsize{0.5} & 48.6 $\pm$ \scriptsize{1.2} & 15.3 $\pm$ \scriptsize{0.6} \cr
DR \citet{hou2018lifelong}
& 59.1 $\pm$ \scriptsize{1.4} & 19.6 $\pm$ \scriptsize{0.5} & 60.8 $\pm$ \scriptsize{1.2} & 17.1 $\pm$ \scriptsize{0.4} & 61.8 $\pm$ \scriptsize{0.9} & 14.3 $\pm$ \scriptsize{0.4} & 46.5 $\pm$ \scriptsize{1.6} & 22.0 $\pm$ \scriptsize{0.5} & 48.7 $\pm$ \scriptsize{1.6} & 18.8 $\pm$ \scriptsize{0.5} & 50.7 $\pm$ \scriptsize{1.2} & 15.1 $\pm$ \scriptsize{0.5} \cr
E2E \citet{castro2018end}
& 60.2 $\pm$ \scriptsize{1.3} & 16.5 $\pm$ \scriptsize{0.5} & 62.6 $\pm$ \scriptsize{1.1} & 12.8 $\pm$ \scriptsize{0.4} & 65.1 $\pm$ \scriptsize{0.8} & 8.9 $\pm$ \scriptsize{0.2} & 47.7 $\pm$ \scriptsize{1.9} & 17.9 $\pm$ \scriptsize{0.4} & 50.8 $\pm$ \scriptsize{1.5} & 13.4 $\pm$ \scriptsize{0.4} & 53.9 $\pm$ \scriptsize{1.2} & 8.8 $\pm$ \scriptsize{0.3} \cr
GD (Ours)
& {\bf 62.1 $\pm$ \scriptsize{1.2}} & {\bf 15.4 $\pm$ \scriptsize{0.4}} & {\bf 65.0 $\pm$ \scriptsize{1.1}} & {\bf 12.1 $\pm$ \scriptsize{0.3}} & {\bf 67.1 $\pm$ \scriptsize{0.9}} & {\bf 8.5 $\pm$ \scriptsize{0.3}} & {\bf 50.0 $\pm$ \scriptsize{1.7}} & {\bf 16.8 $\pm$ \scriptsize{0.4}} & {\bf 53.7 $\pm$ \scriptsize{1.5}} & {\bf 12.8 $\pm$ \scriptsize{0.5}} & {\bf 56.5 $\pm$ \scriptsize{1.2}} & {\bf 8.4 $\pm$ \scriptsize{0.4}} \cr
\hline
\multicolumn{3}{c}{With an external dataset} \cr
LwF \citet{li2016learning}
& 59.7 $\pm$ \scriptsize{1.2} & 19.4 $\pm$ \scriptsize{0.5} & 61.2 $\pm$ \scriptsize{1.1} & 17.0 $\pm$ \scriptsize{0.4} & 60.8 $\pm$ \scriptsize{0.9} & 14.8 $\pm$ \scriptsize{0.4} & 47.2 $\pm$ \scriptsize{1.7} & 21.7 $\pm$ \scriptsize{0.5} & 49.2 $\pm$ \scriptsize{1.3} & 18.6 $\pm$ \scriptsize{0.4} & 49.4 $\pm$ \scriptsize{0.8} & 15.8 $\pm$ \scriptsize{0.4} \cr
DR \citet{hou2018lifelong}
& 59.8 $\pm$ \scriptsize{1.0} & 19.5 $\pm$ \scriptsize{0.5} & 62.0 $\pm$ \scriptsize{0.9} & 16.8 $\pm$ \scriptsize{0.4} & 63.0 $\pm$ \scriptsize{1.0} & 13.9 $\pm$ \scriptsize{0.4} & 47.3 $\pm$ \scriptsize{1.7} & 21.8 $\pm$ \scriptsize{0.6} & 50.2 $\pm$ \scriptsize{1.5} & 18.5 $\pm$ \scriptsize{0.5} & 51.8 $\pm$ \scriptsize{0.9} & 14.9 $\pm$ \scriptsize{0.5} \cr
E2E \citet{castro2018end}
& 61.5 $\pm$ \scriptsize{1.2} & 16.4 $\pm$ \scriptsize{0.5} & 64.3 $\pm$ \scriptsize{1.0} & 12.7 $\pm$ \scriptsize{0.4} & 66.1 $\pm$ \scriptsize{0.9} & 9.2 $\pm$ \scriptsize{0.4} & 49.2 $\pm$ \scriptsize{1.7} & 17.7 $\pm$ \scriptsize{0.6} & 52.8 $\pm$ \scriptsize{1.4} & 13.2 $\pm$ \scriptsize{0.2} & 55.2 $\pm$ \scriptsize{0.9} & 9.0 $\pm$ \scriptsize{0.4} \cr
GD (Ours)
& {\bf 66.3 $\pm$ \scriptsize{1.2}} & {\bf 9.8 $\pm$ \scriptsize{0.3}} & {\bf 68.1 $\pm$ \scriptsize{1.1}} & {\bf 7.7 $\pm$ \scriptsize{0.3}} & {\bf 68.9 $\pm$ \scriptsize{1.0}} & {\bf 5.5 $\pm$ \scriptsize{0.4}} & {\bf 55.2 $\pm$ \scriptsize{1.8}} & {\bf 9.6 $\pm$ \scriptsize{0.4}} & {\bf 57.7 $\pm$ \scriptsize{1.6}} & {\bf 7.4 $\pm$ \scriptsize{0.3}} & {\bf 58.7 $\pm$ \scriptsize{1.2}} & {\bf 5.4 $\pm$ \scriptsize{0.3}} \cr
\bottomrule
\end{tabular}
\cuttabledown
\end{table*}

\begin{figure*}[ht]
\centering
\hspace*{\fill}
\subfigure[ACC improvement by learning with external data]
{\includegraphics[width=\twoawidth]{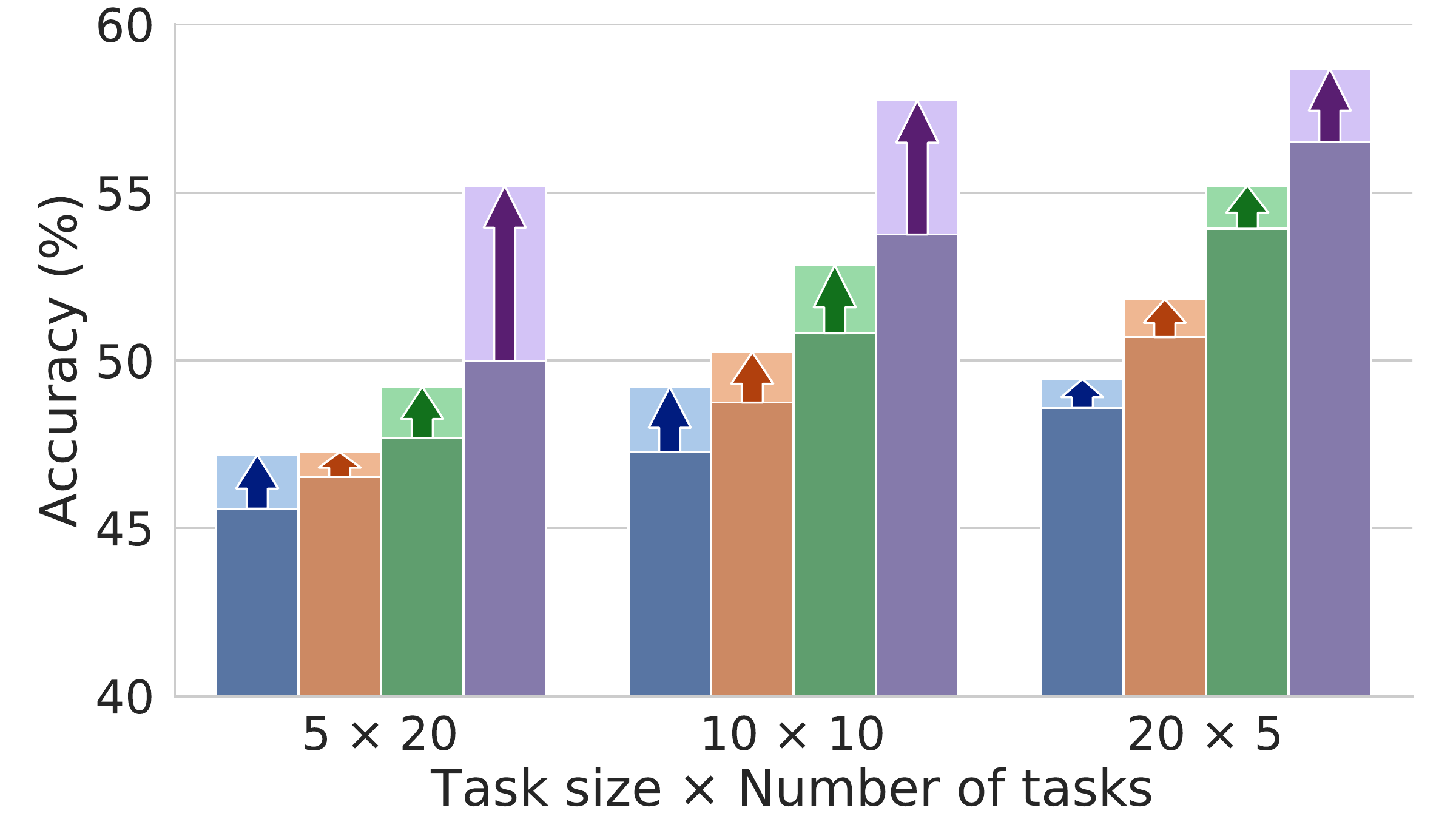}
\label{fig:acc_imagenet}}
\hfill
\subfigure[FGT improvement by learning with external data \hspace{55pt}]
{\includegraphics[width=\twobwidth]{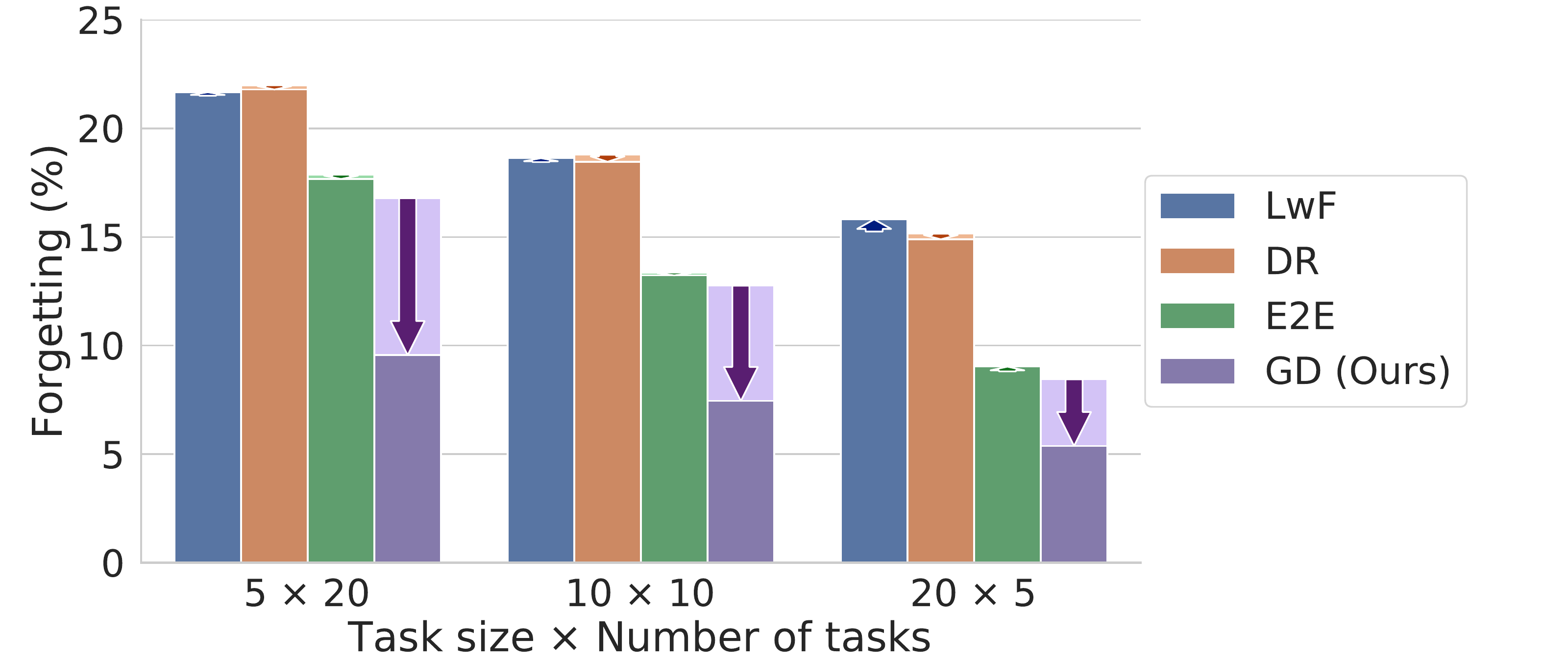}
\label{fig:fgt_imagenet}}
\cutcaptionup
\caption{
Experimental results on ImageNet.
Arrows show the performance gain in ACC and FGT by learning with unlabeled data, respectively.
We report the average performance of nine trials.
}
\cutcaptiondown
\label{fig:exp_imagenet}
\end{figure*}

\begin{figure*}[ht]
\centering
\hspace*{\fill}
\subfigure[ACC with respect to the number of trained classes]
{\includegraphics[width=\twocwidth]{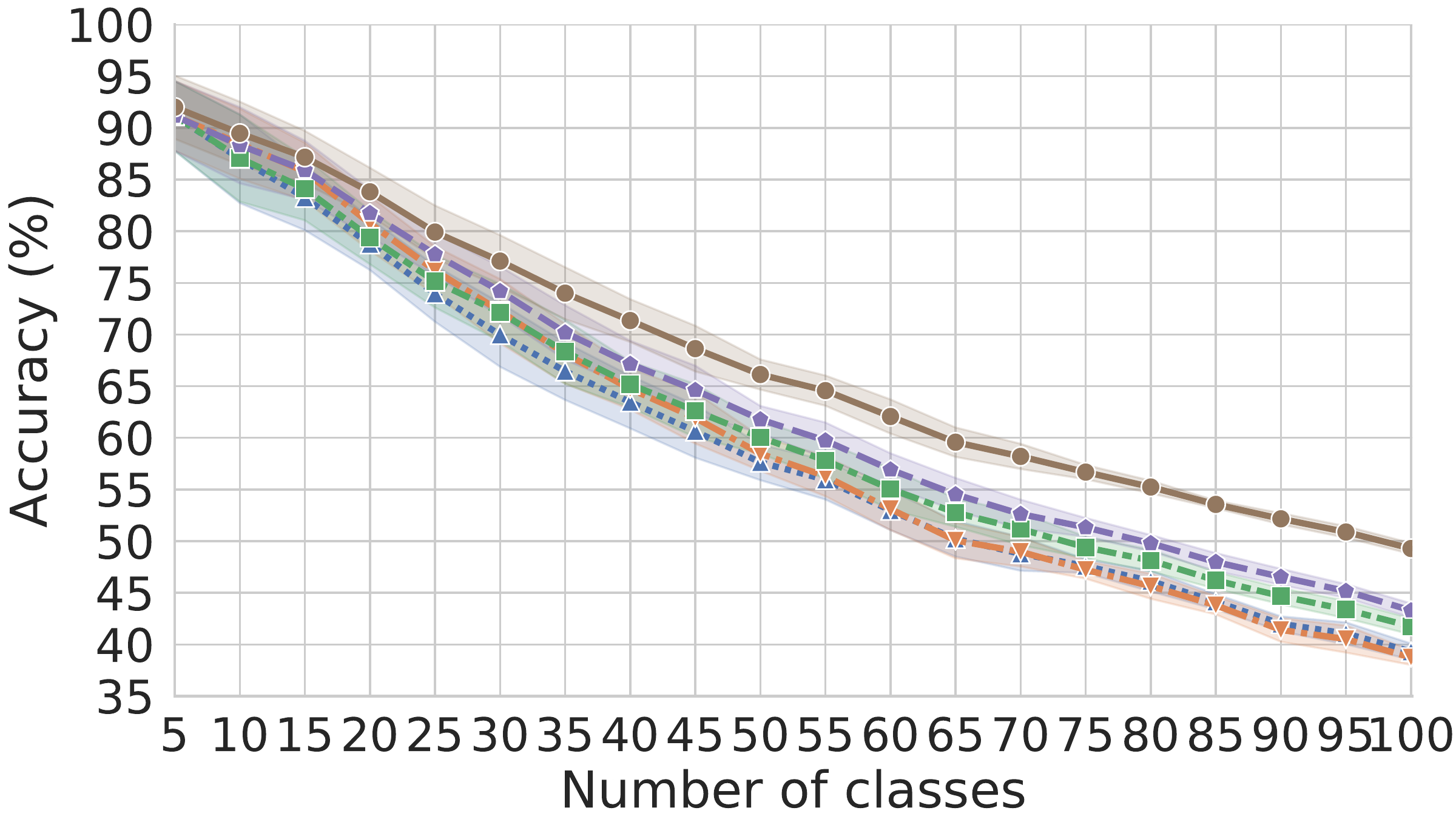}
\label{fig:acc_cifar100_5_5}}
\hfill
\subfigure[FGT with respect to the number of trained classes \hspace{55pt}]
{\includegraphics[width=\twodwidth]{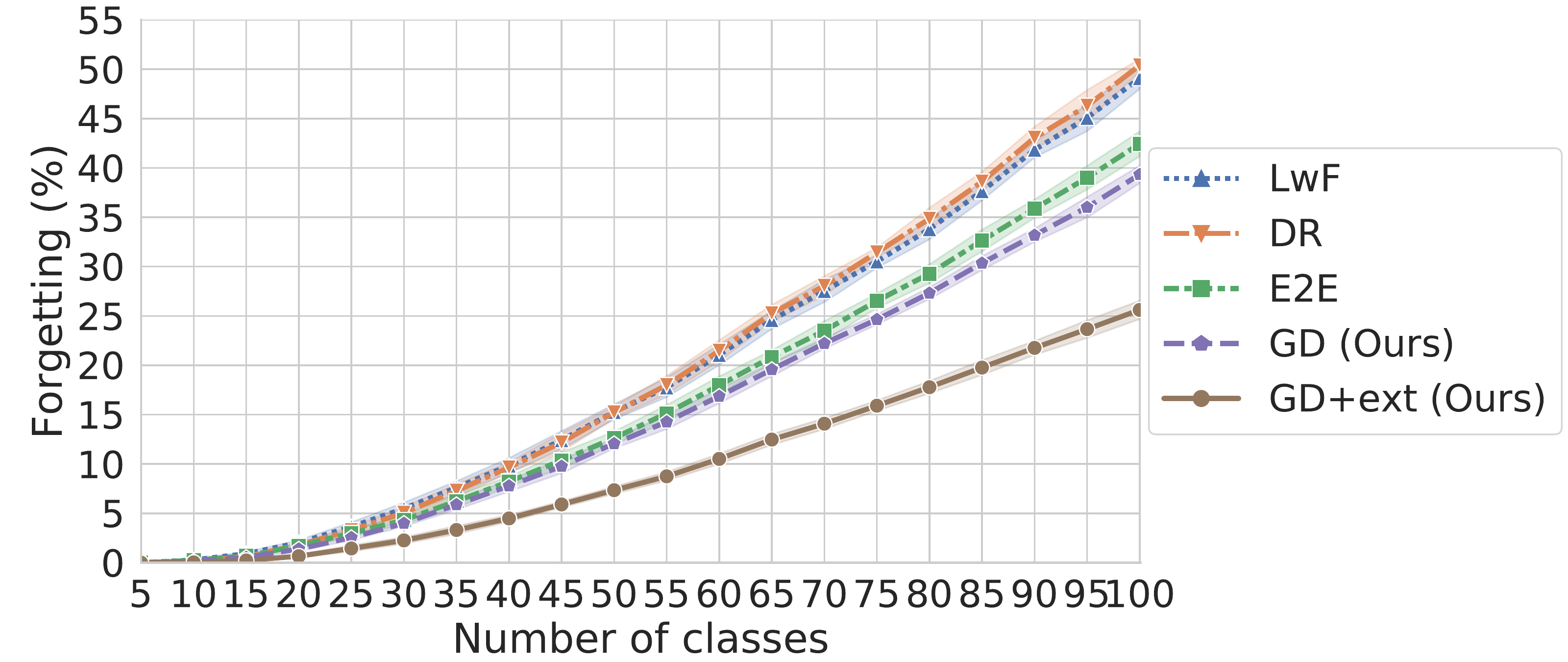}
\label{fig:fgt_cifar100_5_5}}
\cutcaptionup
\caption{
Experimental results on CIFAR-100 when the task size is 5.
We report ACC and FGT with respect to the number of trained classes averaged over ten trials.
}
\cutcaptiondown
\label{fig:exp_cifar100_5_5}
\end{figure*}

\begin{figure*}[ht]
\centering
\hspace*{\fill}
\subfigure[ACC with respect to the number of trained classes]
{\includegraphics[width=\twocwidth]{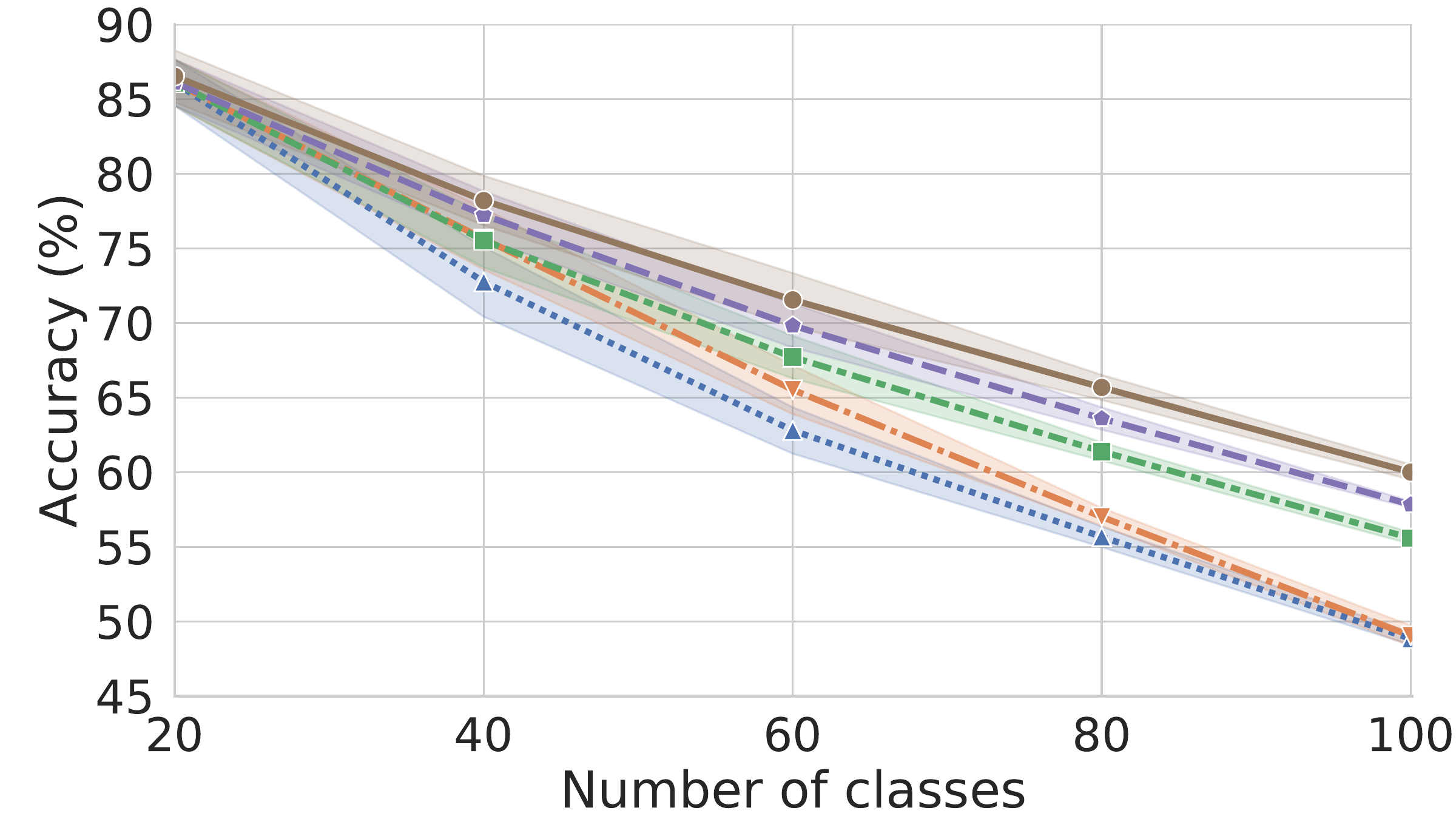}
\label{fig:acc_cifar100_20_20}}
\hfill
\subfigure[FGT with respect to the number of trained classes \hspace{55pt}]
{\includegraphics[width=\twodwidth]{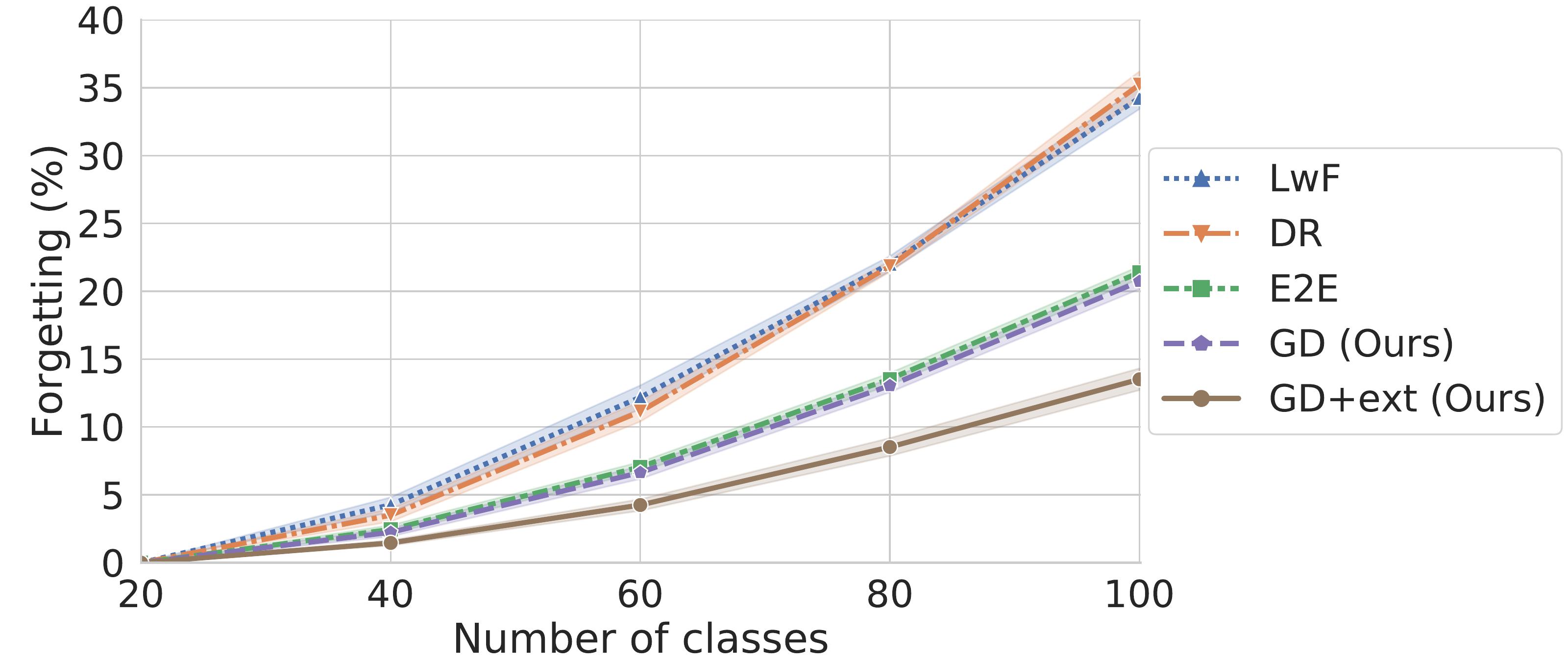}
\label{fig:fgt_cifar100_20_20}}
\cutcaptionup
\caption{
Experimental results on CIFAR-100 when the task size is 20.
We report ACC and FGT with respect to the number of trained classes averaged over ten trials.
}
\cutcaptiondown
\label{fig:exp_cifar100_20_20}
\end{figure*}

\begin{figure*}[ht]
\centering
\hspace*{\fill}
\subfigure[ACC with respect to the number of trained classes]
{\includegraphics[width=\twocwidth]{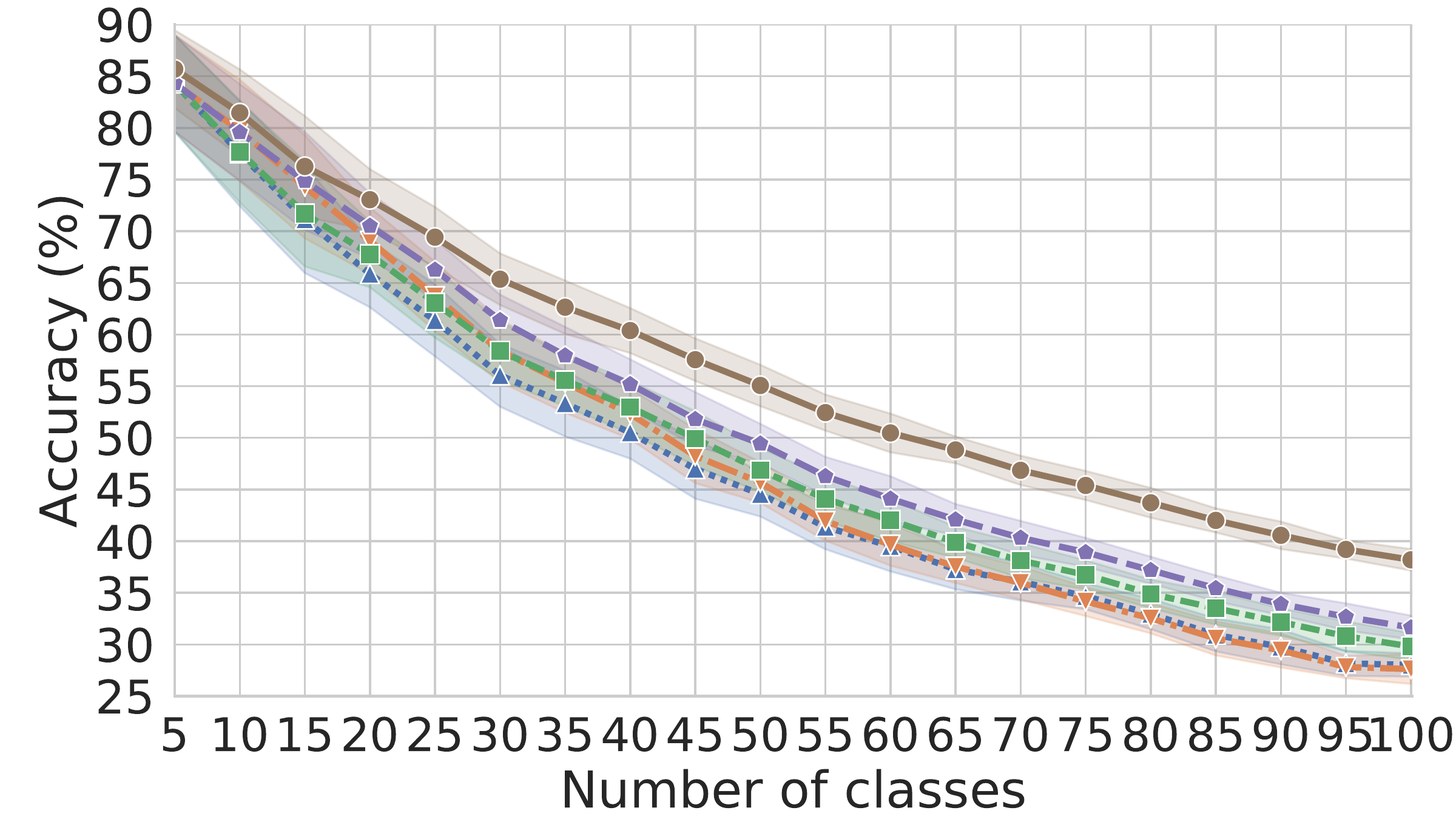}
\label{fig:acc_imagenet_5_5}}
\hfill
\subfigure[FGT with respect to the number of trained classes \hspace{55pt}]
{\includegraphics[width=\twodwidth]{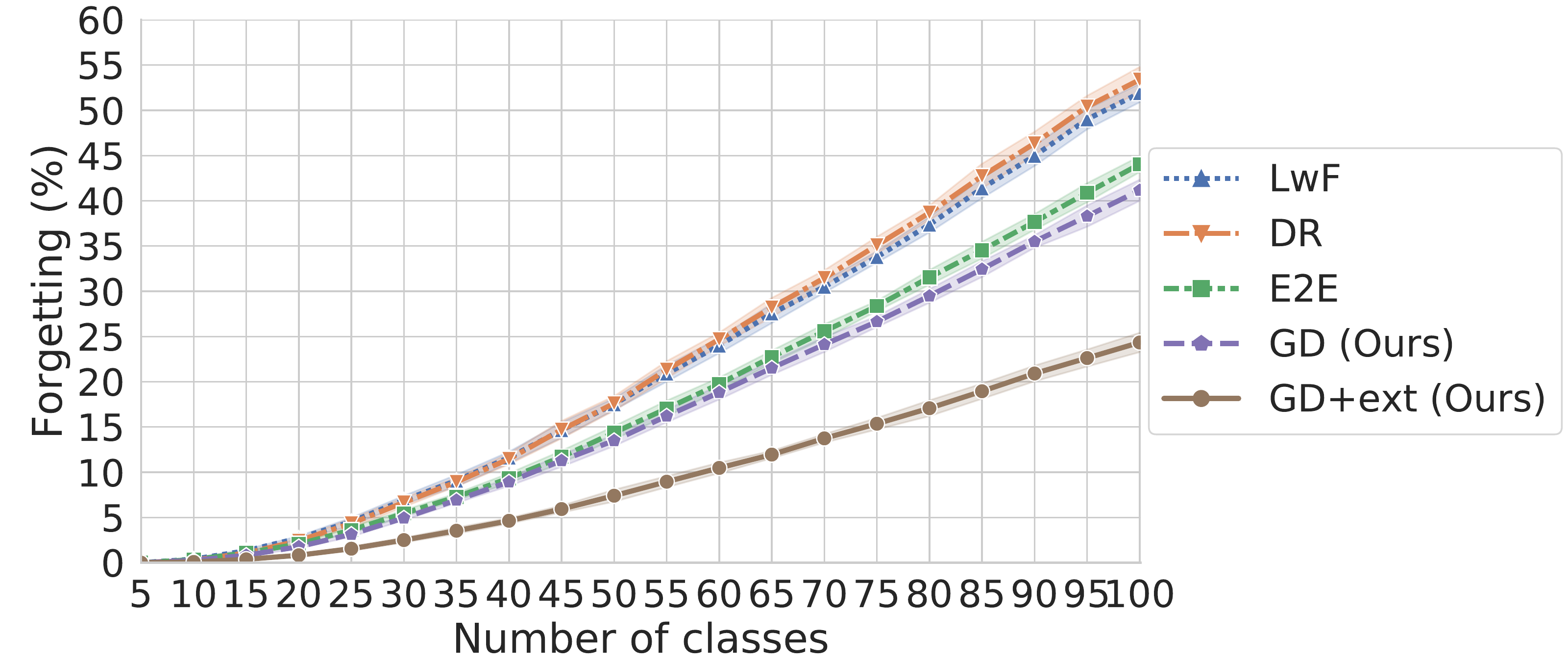}
\label{fig:fgt_imagenet_5_5}}
\cutcaptionup
\caption{
Experimental results on ImageNet when the task size is 5.
We report ACC and FGT with respect to the number of trained classes averaged over nine trials.
}
\cutcaptiondown
\label{fig:exp_imagenet_5_5}
\end{figure*}

\begin{figure*}[ht]
\centering
\hspace*{\fill}
\subfigure[ACC with respect to the number of trained classes]
{\includegraphics[width=\twocwidth]{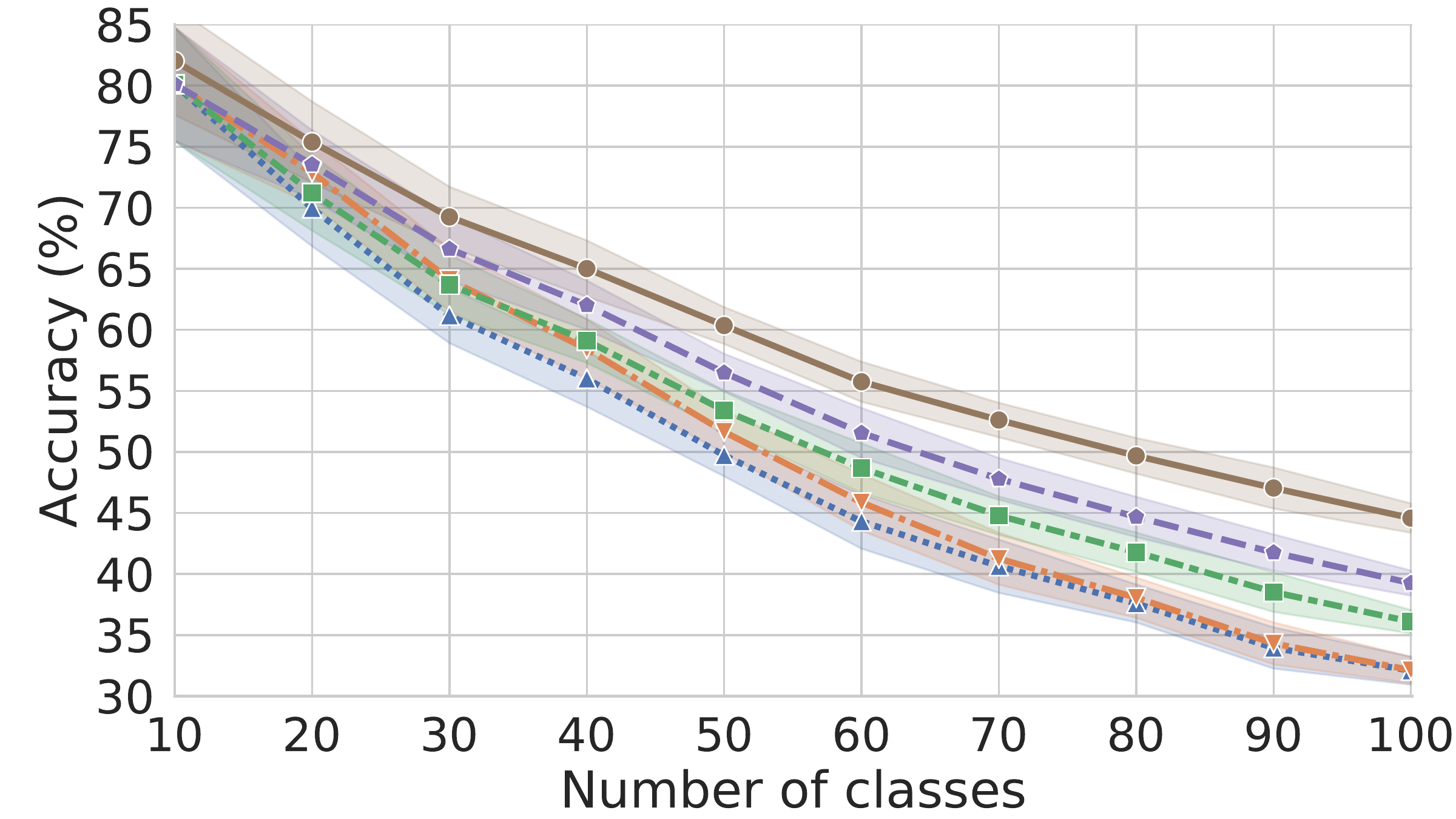}
\label{fig:acc_imagenet_10_10}}
\hfill
\subfigure[FGT with respect to the number of trained classes \hspace{55pt}]
{\includegraphics[width=\twodwidth]{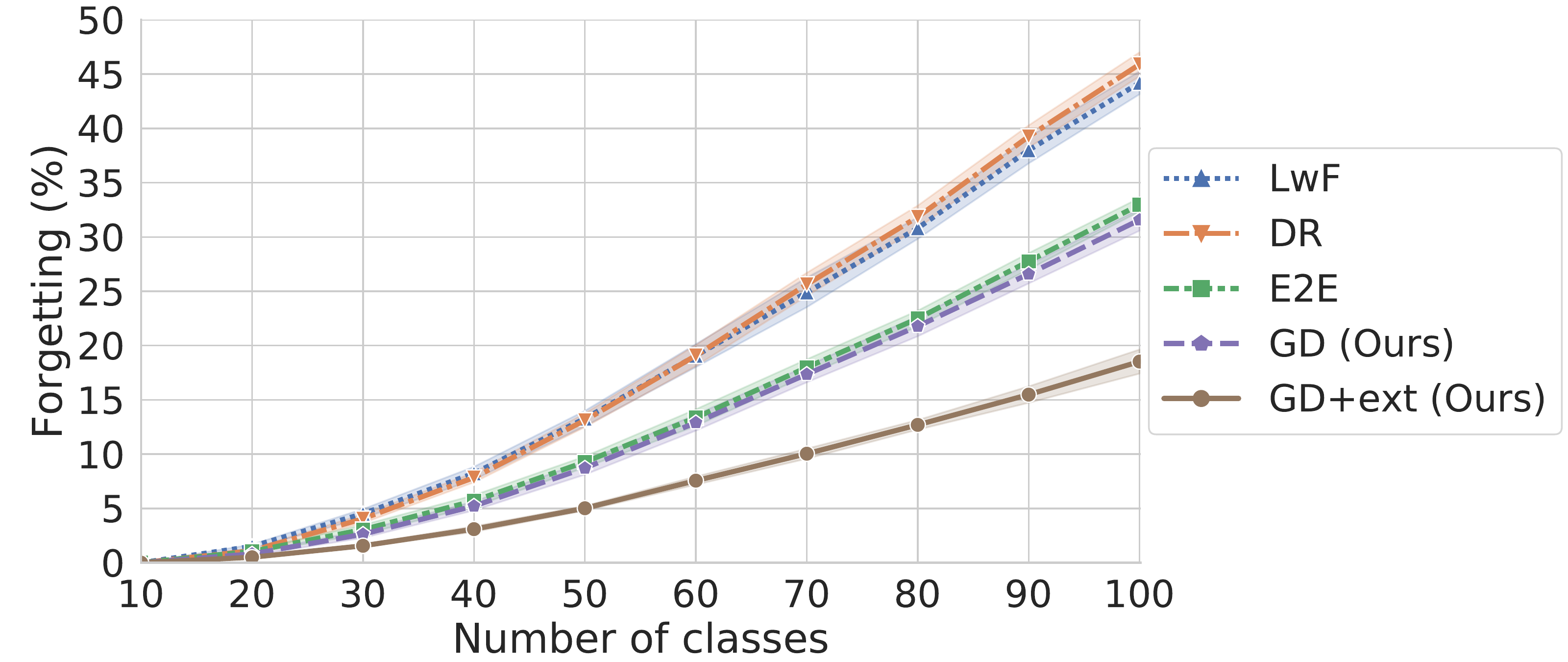}
\label{fig:fgt_imagenet_10_10}}
\cutcaptionup
\caption{
Experimental results on ImageNet when the task size is 10.
We report ACC and FGT with respect to the number of trained classes averaged over nine trials.
}
\cutcaptiondown
\label{fig:exp_imagenet_10_10}
\end{figure*}

\begin{figure*}[ht]
\centering
\hspace*{\fill}
\subfigure[ACC with respect to the number of trained classes]
{\includegraphics[width=\twocwidth]{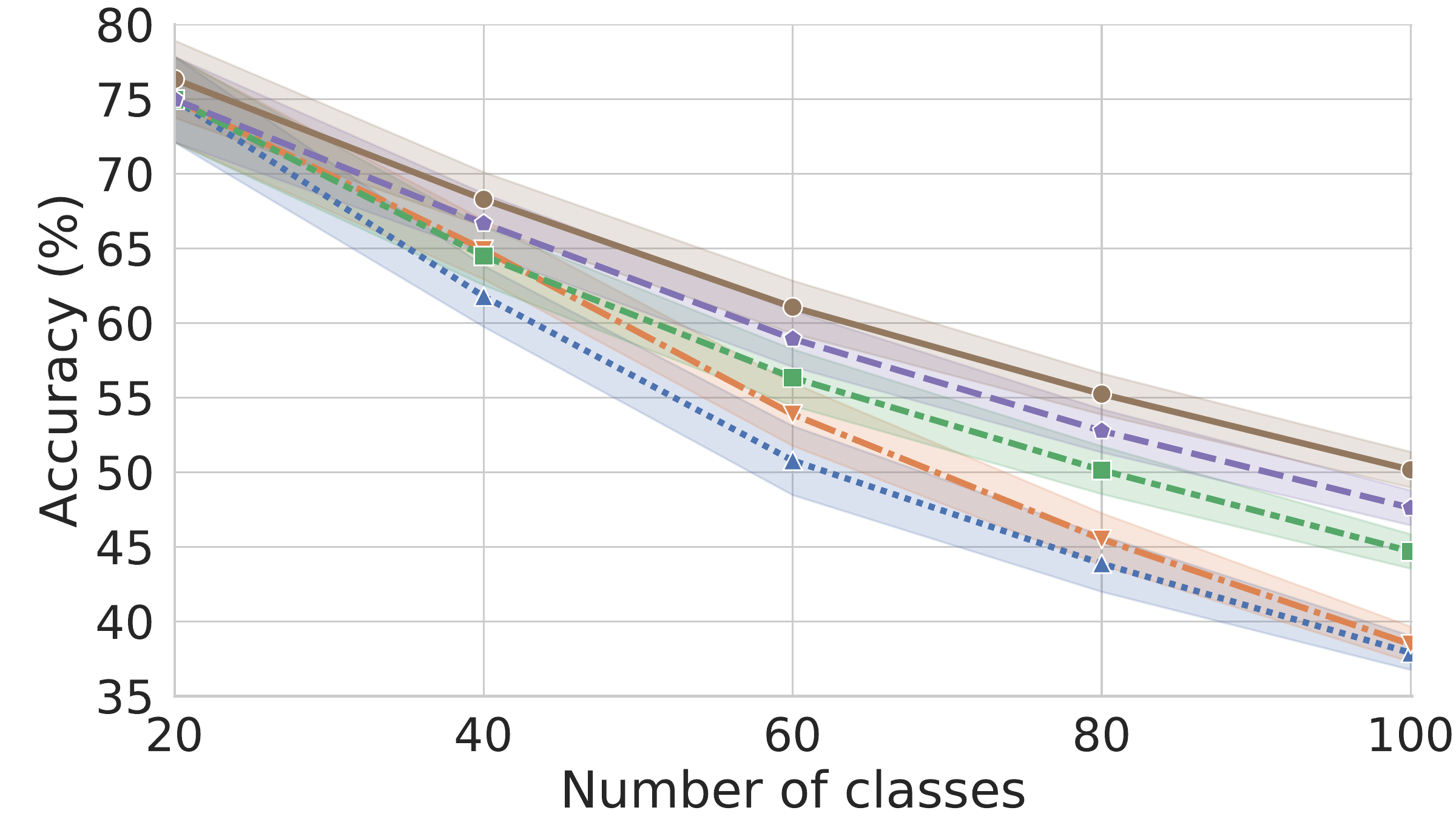}
\label{fig:acc_imagenet_20_20}}
\hfill
\subfigure[FGT with respect to the number of trained classes \hspace{55pt}]
{\includegraphics[width=\twodwidth]{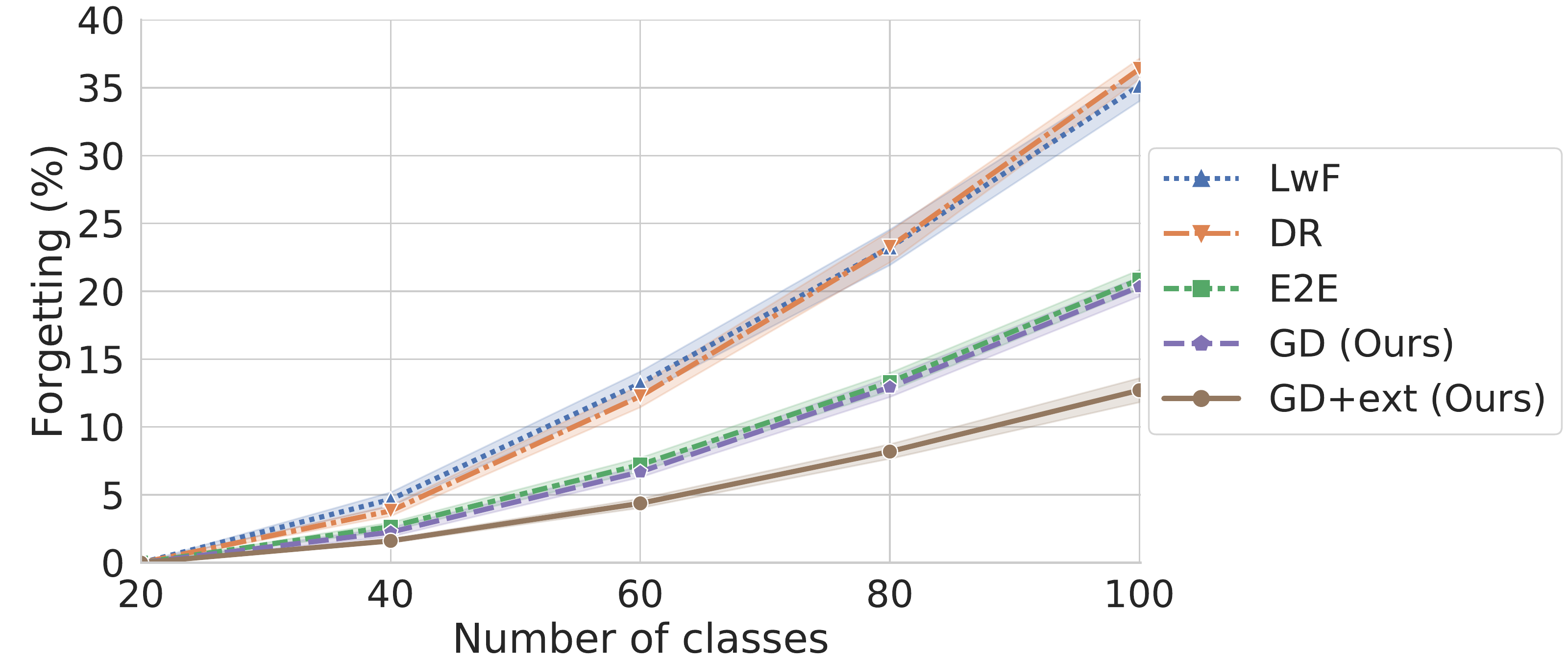}
\label{fig:fgt_imagenet_20_20}}
\cutcaptionup
\caption{
Experimental results on ImageNet when the task size is 20.
We report ACC and FGT with respect to the number of trained classes averaged over nine trials.
}
\cutcaptiondown
\label{fig:exp_imagenet_20_20}
\end{figure*}

\fi

\end{document}